\documentclass{article} %
\usepackage{apex,times}

\usepackage{amsmath,amsfonts,bm}

\def\eqref#1{equation~\ref{#1}}

\def\1{\bm{1}}

\DeclareMathAlphabet{\mathsfit}{\encodingdefault}{\sfdefault}{m}{sl}
\SetMathAlphabet{\mathsfit}{bold}{\encodingdefault}{\sfdefault}{bx}{n}

\usepackage{graphicx}
\usepackage{wrapfig}
\usepackage{needspace}
\usepackage{caption}
\newcommand{\apexwrapbegin}{}
\newcommand{\apexwrapend}{\WFclear}
\usepackage{booktabs}
\usepackage{tabularx}
\usepackage{multirow}
\usepackage{xcolor}
\usepackage{algorithm}
\usepackage{algpseudocode}
\usepackage{listings}
\usepackage{tikz}
\usepackage{pgfplots}
\pgfplotsset{compat=1.18}
\usepackage{hyperref}
\usepackage{url}
\hypersetup{
  colorlinks=true,
  citecolor=black,
  linkcolor=black,
  urlcolor=blue
}

\lstdefinestyle{apexcommand}{
  basicstyle=\ttfamily\scriptsize,
  columns=fullflexible,
  keepspaces=true,
  breaklines=true,
}

\usepackage{amsmath,amssymb,graphicx,booktabs,array,tabularx,longtable}
\usepackage{xcolor,listings,caption,pgfplots,url,needspace}
\pgfplotsset{compat=1.18}
\definecolor{apexgreen}{RGB}{0,112,62}
\definecolor{apexmissing}{RGB}{175,30,30}
\definecolor{apexgray}{gray}{0.95}

\providecommand{\apexSquare}[3]{%
  \IfFileExists{#1}{\includegraphics[width=#2,height=#2,keepaspectratio]{#1}}{%
  \begingroup\setlength{\fboxsep}{0pt}\colorbox{apexgray}{%
  \parbox[c][#2][c]{#2}{\centering\scriptsize\color{gray}#3}}\endgroup}}
\providecommand{\apexWideSlot}[3]{%
  \IfFileExists{#1}{\includegraphics[width=\linewidth,height=#2,keepaspectratio]{#1}}{%
  \begingroup\setlength{\fboxsep}{5pt}\fcolorbox{gray}{apexgray}{%
  \parbox[c][#2][c]{\dimexpr\linewidth-2\fboxsep-2\fboxrule\relax}{\centering\small\color{gray}#3}}\endgroup}}
\lstdefinestyle{apexpseudo}{
  basicstyle=\ttfamily\footnotesize,columns=fullflexible,keepspaces=true,
  showstringspaces=false,breaklines=true,breakatwhitespace=false,
  aboveskip=5pt,belowskip=5pt,frame=none,
  moredelim=**[is][\color{apexgreen}]{@@}{@@}}
\newcounter{apexalgorithm}
\newcommand{\apexAlgCaption}[2]{\refstepcounter{apexalgorithm}\label{#2}%
  \par\noindent\textbf{Algorithm \theapexalgorithm. #1}\par}

\newcommand{\dgfid}[1]{\textcolor[rgb]{0.75,0.15,0.12}{$+#1$}}

\title{\texorpdfstring{APEX: Asynchronous Prefix Denoising\\ for Extreme Compression}{APEX: Asynchronous Prefix Denoising for Extreme Compression}}

\author{%
  David Dinkevich$^{1,2}$ \quad
  Nisan Chiprut$^{1}$ \quad
  Yoav HaCohen$^{1}$ \quad
  Dani Lischinski$^{2}$ \\[0.4em]
  $^{1}$Lightricks \quad $^{2}$Hebrew University of Jerusalem
}

\fancypagestyle{apexfirst}{%
  \fancyhead{}%
  \renewcommand{\headrulewidth}{0pt}%
}

\begin{document}

\maketitle
\lhead{\small\itshape APEX: Asynchronous Prefix Denoising for Extreme Compression}
\thispagestyle{apexfirst}

\begin{abstract}
Fewer latent tokens promise cheaper diffusion. Yet aggressive spatial compression creates a reconstruction-generation dilemma: adding channels improves reconstruction but makes generation harder. Our goal is to make extreme spatial compression competitive by shifting this frontier. We introduce APEX, short for Asynchronous Prefix denoising for EXtreme compression, which pairs a hierarchical latent space with asynchronous denoising over channel groups. Our autoencoder orders channel prefixes from structure to detail, while one shared denoiser generates them at staggered local times without adding tokens. We show that naively combining a channel hierarchy with asynchronous denoising can make generation worse. APEX succeeds by matching what each level represents, when it denoises, and how training is allocated across the trajectory. On ImageNet-512, APEX outperforms existing methods at every evaluated channel count under \(64\times\) and \(128\times\) spatial compression, substantially improving this frontier. At matched training and guided sampling compute, it also outperforms a 256-token model at every evaluated budget, with larger gains at lower compute. APEX makes extreme spatial compression practical for compute-efficient latent diffusion.
Code is available at \url{https://daviddinkevich.github.io/APEX}.
\end{abstract}

\section{Introduction}

Modern diffusion models increasingly use Transformers, whose attention cost grows quadratically with the number of input tokens. Latent diffusion reduces this cost by compressing an image into a smaller spatial grid and applying diffusion there \citep{rombach2022ldm}. Our goal is to make extreme spatial compression competitive for latent diffusion. Stronger spatial compression produces fewer tokens. Channel count is a separate design choice, but preserving reconstruction quality typically requires adding channels as the grid shrinks, producing increasingly \emph{wide} latents. This trade is attractive because shorter sequences can substantially reduce denoiser cost, permitting more optimization and sampling under a fixed compute budget. This suggests an obvious strategy: compress the spatial grid as aggressively as possible, then add enough channels to preserve reconstruction quality.

Unfortunately, this strategy encounters a well-established reconstruction-generation dilemma \citep{yao2025vavae,chen2025dcae,chen2025dcae15,lai2025freqwarm}. At fixed spatial compression, adding channels improves reconstruction but makes the resulting latent harder to generate. Additional channels let the encoder preserve more image-specific information, including exact textures and local details. This lowers reconstruction error, but increases the information that the diffusion model must produce \citep{chen2025dcae15,heek2026unifiedlatents}. Standard latent diffusion therefore faces a poor choice: use fewer channels and sacrifice reconstruction quality, or preserve more detail and sacrifice generation quality. Existing approaches reduce this tension by restricting latent information or importing pretrained semantic structure \citep{kingma2013vae,yu2025repa,yao2025vavae,zheng2025rae,pan2025sfd}. The closest structured-latent method in our setting organizes channel prefixes within one reconstruction latent, but still denoises each retained prefix at a shared timestep \citep{chen2025dcae15}. This limitation becomes especially severe at \(128\times\) spatial compression, where a \(512^2\) image is represented by a \(4\times4\) grid of only 16 tokens.

\begin{figure}[t]
\begin{center}
\includegraphics[width=\linewidth]{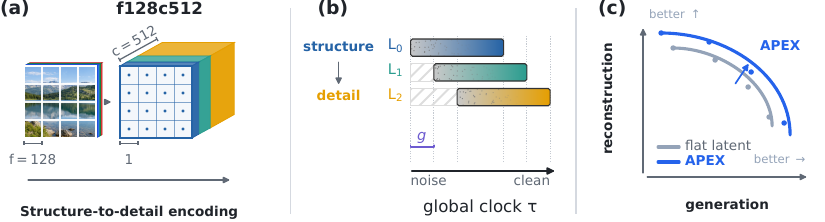}
\end{center}
\caption{\textbf{APEX enables extreme-compression latent diffusion.}
\textbf{(a)} An f128c512 latent represents a \(512^2\) image with 16 spatial tokens and orders channels from structure to detail.
\textbf{(b)} One global clock denoises those levels at staggered local times; \(g\) is the delay before each later level starts.
\textbf{(c)} Ordering generation this way improves the reconstruction--generation trade-off relative to a flat latent.}
\label{fig:teaser}
\end{figure}

APEX addresses this problem without imposing an additional bottleneck on the final latent. Instead, it changes how the channels are organized and generated. Coarse structure must be generated from scratch, while exact texture and local detail become easier to predict once the scene is established. We therefore factorize generation into structure generation followed by conditional refinement. Our autoencoder orders channel prefixes from structure to detail by supervising them at progressively higher resolutions. During diffusion, one shared Transformer denoises the newly added channel groups at staggered local times, so each refinement is predicted from cleaner preceding context. As this is done along the channel dimension, APEX does not increase the number of tokens. Figure~\ref{fig:teaser} summarizes the representation and denoising process. Although hierarchical channel latents and asynchronous denoising each exist, combining them is not automatic: shared-time diffusion performs poorly on APEX levels, while staggered local times perform poorly with the DC-AE 1.5 hierarchy and prefix sampler (Table~\ref{tab:component_ablation}).

We evaluate APEX on ImageNet-512 at \(64\times\) and \(128\times\) spatial compression. We write \(f64c128\) for \(64\times\) spatial compression with 128 channels. At \(512^2\), \(f64\) and \(f128\) correspond to 64 and 16 tokens. Across every evaluated channel count, APEX outperforms all compared methods and degrades far less as channels are added. At \(f128c512\), APEX reaches \(8.98\) guided gFID from 16 tokens with SiT-XL, while every compared method remains in the double digits. Our 64-token \(f64c128\) model matches the gFID of the widely used 1024-token \(f8c4\) SD-VAE \citep{rombach2022ldm} using \(15\%\) of its denoiser-training FLOPs (Appendix~\ref{app:sdvae}). With training compute matched, our 64-token hierarchical model also outperforms a less-compressed 256-token model across every evaluated guided inference budget, with larger gains as sampling compute decreases. These results show that once the reconstruction-generation dilemma is addressed, extreme spatial compression becomes a quality-compute advantage rather than only a per-step speedup.

Our contributions are threefold:

\begin{enumerate}
\item \textbf{Making hierarchy and asynchronous denoising work together.}
We show that nested channel levels and staggered local times do not
combine automatically. APEX couples resolution-supervised levels,
local times, and training allocation within one fixed-token latent.
\item \textbf{Turning high compression into a compute advantage.}
With training compute matched, APEX at 64 tokens outperforms a less-compressed 256-token baseline across every evaluated guided inference budget, with larger gains at lower compute.
\item \textbf{Generating without an additional information bottleneck.} Rather than improving generative tractability by restricting what the latent can store, APEX improves generative tractability while maintaining comparable reconstruction quality, without imposing an additional explicit information bottleneck.
\end{enumerate}

\section{Related Work}
\label{sec:related}

\paragraph{Deep Compression Autoencoders.}
Deep-compression autoencoders reduce diffusion sequence length by
moving representational capacity from spatial positions into latent
channels. DC-AE made reconstruction practical at \(32\times\),
\(64\times\), and \(128\times\) spatial compression and provides our
base architecture \citep{chen2025dcae}. DC-AE 1.5 structures channel
prefixes and trains diffusion on randomly sampled prefixes at a shared
timestep, while FreqWarm modifies diffusion training through a
frequency curriculum \citep{chen2025dcae15,lai2025freqwarm}. These are
our closest direct baselines. APEX instead pairs resolution-matched
cumulative levels with staggered local times; APEX levels under
shared-time diffusion and a direct application of the staggered schedule to the DC-AE 1.5 construction perform poorly
(Table~\ref{tab:component_ablation}).

\paragraph{Latent Information and Semantic Alignment.}
Existing methods improve diffusibility either by controlling latent
information or importing pretrained semantic structure. Traditional
VAEs impose a Gaussian prior, Unified Latents explicitly controls
information rate with a learned diffusion prior, and MAETok uses a
non-variational discriminative latent for 128-token generation
\citep{kingma2013vae,heek2026unifiedlatents,chen2025maetok}.
Semantic-alignment methods instead shape the latent or denoiser using
pretrained representations
\citep{yu2025repa,yao2025vavae,zheng2025rae,
kouzelis2025redi,wu2025reg,leng2025repae}.
SFD and Learning When to Denoise additionally pair a DINOv2-derived
semantic latent with a separate texture latent and denoise semantics
first using a fixed or learned relative schedule
\citep{pan2025sfd,qian2026lwd,oquab2024dinov2}.
APEX uses no external semantic representation or explicit latent
prior. It induces structure within one reconstruction latent, leaving
later levels free to preserve image-specific detail.

\paragraph{Hierarchical and Frequency-Based Representations.}
Coarse-to-fine representations appear in Matryoshka learning,
cascaded and nested diffusion, multiscale tokenizers, ordered-channel
models, and flexible-length token sequences
\citep{kusupati2022matryoshka,ho2022cascaded,gu2024matryoshka,
chen2025hieratok,zhong2025dcsldm,paul2026channeltok,
cai2026davae,bachmann2025flextok}.
Frequency-based methods instead reshape latent spectra or explicitly
separate coarse and fine components
\citep{ning2026spectrum,skorokhodov2025scale,
kouzelis2025eqvae,li2025wfvae,guan2026lcvae}.
Denoising-VAE constructs progressively low-pass latent levels with
blurred image targets and frequency-aligned diffusion
\citep{xiang2025denoisingvae}; we use Spectrum Matching as the
representative autoencoder-side baseline for this family. WaiT
similarly delays wavelet-detail generation, but operates in pixel
space rather than a learned high-compression latent
\citep{pavasovic2026wait}. APEX instead uses cumulative channel
prefixes of one jointly decoded latent on a fixed spatial grid,
supervising each prefix at a corresponding resolution and generating
later groups from cleaner preceding levels.

\section{Method}
\label{sec:method}

\begin{figure}[t]
\begin{center}
\includegraphics[width=\linewidth]{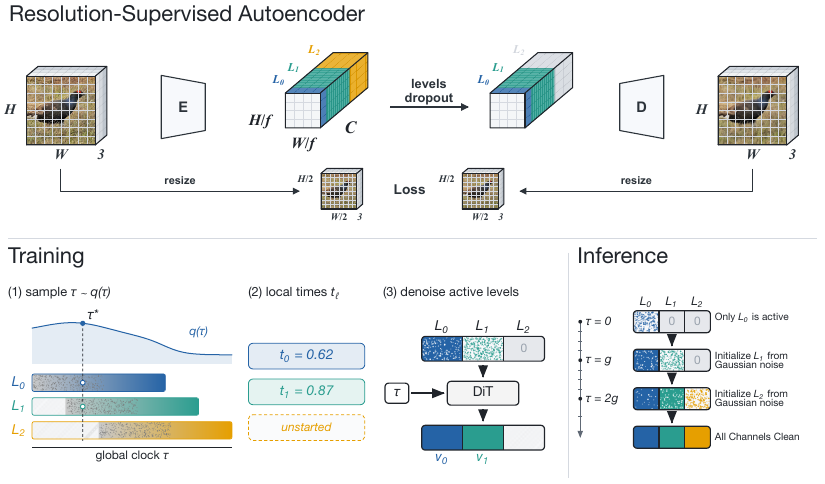}
\end{center}
\caption{\textbf{Channel hierarchy.}
We divide the channels of each latent token into ordered groups.
The cumulative prefix through group $\ell$ defines level $L_\ell$.
\textbf{Resolution-supervised autoencoder.}
We sample a level, zero all later groups, and decode the retained prefix at full resolution.
Reconstruction and target are then resized to that level's assigned resolution before computing the autoencoder loss.
\textbf{Diffusion training.}
A global clock $\tau$ is sampled from $q(\tau)$ and mapped to one local timestep per group.
Unstarted groups are zeroed, and the shared DiT predicts velocities for the denoising groups.
\textbf{Inference.}
Groups start in sequence along the same clock.
Each newly active group is initialized from Gaussian noise, and generation continues until the full latent is clean.}
\label{fig:ae_method}
\end{figure}

APEX has two components. First, a single autoencoder orders cumulative channel levels from structure to detail using progressively higher-resolution targets (Section~\ref{sec:method_ae}). Second, a shared Transformer denoises the channels introduced at each level at staggered local times, allowing later levels to refine cleaner earlier ones without adding spatial tokens (Section~\ref{sec:method_diffusion}; Figure~\ref{fig:ae_method}).
Pseudocode for autoencoder training, diffusion training, and inference is given in Appendices~\ref{app:ae_training} and~\ref{app:sampling} (Algorithms~\ref{alg:apex_ae}, \ref{alg:apex_train}, and~\ref{alg:sampling}).
\subsection{Resolution-Aware Autoencoder Training}
\label{sec:method_ae}

Given an image \(x\in\mathbb{R}^{R\times R\times3}\),
the encoder produces a latent
\(z\in\mathbb{R}^{(R/f)\times(R/f)\times C}\), which we partition as
\begin{equation}
z=[b_0\Vert b_1\Vert\cdots\Vert b_{L-1}],
\qquad
b_\ell\in
\mathbb{R}^{(R/f)\times(R/f)\times d_\ell}.
\label{eq:channel_blocks}
\end{equation}
We call the cumulative representation a level:
\begin{equation}
z^{(\ell)}=[b_0\Vert\cdots\Vert b_\ell]
\in
\mathbb{R}^{(R/f)\times(R/f)\times C_\ell},
\qquad
C_\ell=\sum_{k=0}^{\ell}d_k,
\label{eq:cumulative_level}
\end{equation}
It retains all preceding channels while adding capacity on the same spatial
grid.

We use a simple structure-to-detail heuristic to choose target resolutions. At lower channel counts, a roughly \(4\times\) increase in cumulative channels corresponds to a \(2\times\) increase in target resolution along each spatial dimension. At larger widths, we treat this proportionality as a guideline rather than a constraint: we introduce intermediate channel levels to avoid excessively large refinement groups and use monotonically increasing target resolutions up to the full image resolution. Exact assignments are given in Appendix Table~\ref{tab:app_hierarchies}. Any additional levels are also supervised at full resolution. We denote a hierarchy by its spatial compression followed by its cumulative level sizes; for example, f64c\(8{-}32{-}128\) has \(64\times\) spatial compression and levels with 8, 32, and 128 cumulative channels, assigned to \(128^2\), \(256^2\), and \(512^2\) targets, respectively.
At each training step, we sample a level uniformly,
\begin{equation*}
    \ell\sim\operatorname{Unif}\{0,\ldots,L-1\},
\end{equation*}
retain its cumulative representation \(z^{(\ell)}\), and set all later channel blocks to zero. The shared decoder produces
\begin{equation}
\hat{x}^{(\ell)}
=
D\!\left(
[b_0\Vert\cdots\Vert b_\ell\Vert0\Vert\cdots\Vert0]
\right).
\label{eq:partial_reconstruction}
\end{equation}
Although the decoder always outputs the original image resolution, we resize both the reconstruction and target to \(R_\ell\times R_\ell\) before computing the loss:
\begin{equation}
\mathcal{L}_{\mathrm{AE}}
=
\mathbb{E}_{x,\ell}
\left[
\mathcal{L}_{\mathrm{base}}
\left(
P_\ell(\hat{x}^{(\ell)}),
P_\ell(x)
\right)
\right],
\label{eq:ae_objective}
\end{equation}
where \(P_\ell\) is an antialiased resize and \(\mathcal{L}_{\mathrm{base}}\) follows the reconstruction, perceptual, and adversarial objective of the underlying DC-AE \citep{chen2025dcae}. Exact channel counts, target resolutions, and training details are given in Section~\ref{sec:experimental_setup} and Appendix~\ref{app:ae_training}.

Resolution-aware supervision discourages narrow levels from spending their limited capacity on isolated high-frequency features. When every level is supervised at full resolution, early reconstructions often preserve sharp edges without recovering the surrounding structure. Resizing the loss instead encourages early levels to preserve coherent layout and shape, while later levels restore the missing detail (Figure~\ref{fig:prefix_reconstructions}).

\Needspace{3.2in}
\vspace*{-12pt}
\subsection{Staggered Channel Diffusion}
\label{sec:method_diffusion}
\csname @nobreakfalse\endcsname
\apexwrapbegin
\setlength{\intextsep}{0pt}
\begin{wrapfigure}[14]{r}{0.48\linewidth}
\vspace{-40pt}
\centering
\includegraphics[width=\linewidth]{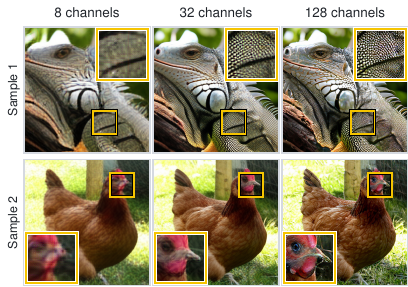}
\captionsetup{justification=raggedright,singlelinecheck=false,font=footnotesize,skip=2pt}
\caption{\textbf{Staged generation.} Decodes after each level reaches \(t=0\). Additional samples in Appendix~\ref{app:qual_generations}.}
\label{fig:staged_generation}
\vspace{\baselineskip}
\end{wrapfigure}
\setlength{\intextsep}{12pt plus 2pt minus 2pt}
During diffusion, each block receives its own local timestep
while remaining on the same spatial grid.

\paragraph{A global clock and level states.}
A stagger schedule coordinates the levels through a single
global clock \(\tau\): each level carries a local time
\(t_\ell(\tau)\) that decreases monotonically from \(1\) (noise)
to \(0\) (clean) over its own window of the clock, and the
windows are ordered so that
\begin{equation}
t_0(\tau)\;\leq\;t_1(\tau)\;\leq\;\cdots\;\leq\;t_{L-1}(\tau).
\label{eq:ordered_times}
\end{equation}
\apexwrapend
Earlier levels are therefore never noisier than the refinement
levels they condition. At any clock value, each level is in one
of three states: \emph{unstarted} (its window has not begun),
\emph{denoising} (\(0<t_\ell(\tau)<1\)), or \emph{done}
(\(t_\ell(\tau)=0\)). Unstarted blocks are zeroed in the
Transformer input; denoising blocks are noised at their local
times; done blocks appear clean and act as fixed context. The
shared network predicts velocities for all blocks, but only
denoising levels are supervised and updated. Predictions on
unstarted and done blocks are discarded.

\paragraph{Equal-gap stagger.}
Our main schedule delays consecutive levels by a constant gap
\(g\):
\begin{equation}
t_\ell(\tau)
=
\operatorname{clip}
\bigl(
1+\ell\cdot g-\tau,\,0,\,1
\bigr),
\qquad
\tau\in[0,\,1+(L-1)g],
\label{eq:stagger_clock}
\end{equation}
so level \(\ell\) is unstarted for \(\tau\leq\ell\cdot g\) (a multiple of the gap), denoising
for \(\ell\cdot g<\tau<1+\ell\cdot g\), and done afterward: the first level
begins immediately, each later level starts after an additional
delay \(g\), and earlier levels are always further along their
trajectories, providing cleaner context for refinement. The
formulation is schedule-agnostic, and any monotone local clocks
satisfying Eq.~\ref{eq:ordered_times} are valid. We adopt the equal-gap
form throughout for its simplicity.

\paragraph{Training.}
We draw the clock from a sampling density \(\tau\sim q\) and noise every
denoising block with the standard flow-matching interpolation
\citep{lipman2023flowmatching,liu2022rectifiedflow}, each at its own local time:
\begin{equation}
\tilde{b}_\ell
=
(1-t_\ell)\,b_\ell+t_\ell\,\epsilon_\ell,
\qquad
v_\ell
=
\epsilon_\ell-b_\ell,
\qquad
\epsilon_\ell\sim\mathcal{N}(0,I).
\label{eq:block_interpolation}
\end{equation}
The objective weights the per-level velocity errors,
\begin{equation}
\mathcal{L}(\tau)
=
\sum_{\ell}
w_\ell(\tau)\,
\bigl\|\hat{v}_\ell-v_\ell\bigr\|^2,
\label{eq:loss}
\end{equation}
where the error of each level is averaged over its spatial
positions and channels. We set \(w_\ell(\tau)\) to zero outside
the denoising state and constant across denoising levels
(normalized to sum to one), which prevents wide refinement
levels from dominating training solely because they contain
more channels.

\apexwrapbegin
\begin{wrapfigure}{r}{0.22\linewidth}
\centering
\includegraphics[width=\linewidth]{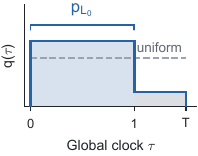}
\captionsetup{font=footnotesize,labelfont=rm}
\caption{Global clock density.}
\label{fig:clock_density}
\end{wrapfigure}
The schedule fixes the levels' relative progress; \(q\) decides
how often each part of the trajectory is trained
(Figure~\ref{fig:clock_density}).
Because \(L_0\) must learn structure from scratch, we assign probability \(p_{L_0}\) to \(0<\tau<1\), where \(L_0\) is denoising, and spread the remaining probability uniformly over the rest of the clock.
A sample from this interval also trains every later level whose window has already opened.
Section~\ref{sec:ablations} analyzes the effect of this sampling mechanism.

\paragraph{Sampling.}
Inference discretizes the clock into a grid
\(0=\tau_0<\cdots<\tau_N\) that includes each activation point
\(\ell\cdot g\) (multiples of the gap).
\apexwrapend
At every step, one shared network evaluation serves
all levels, and each block advances by its own time increment:
\begin{equation}
\tilde{b}_\ell
\;\leftarrow\;
\tilde{b}_\ell
+
\bigl(t_\ell(\tau_{k+1})-t_\ell(\tau_k)\bigr)\,\hat{v}_\ell.
\label{eq:sampler_update}
\end{equation}
Unstarted and done levels have constant local time, so their
increments vanish and they do not move. When a level's window
opens, its block is initialized from Gaussian noise
(Algorithm~\ref{alg:sampling}, Appendix~\ref{app:sampling}).
Under the equal-gap schedule the
increment is \(-\Delta\tau\) for every denoising level and zero
otherwise, so each step reduces to a single shared Euler update
over the denoising blocks: one network evaluation per step, the
same cost per step as a flat model. A nonlinear clock changes
only the increments in Eq.~\ref{eq:sampler_update}.
When each level reaches \(t=0\), we decode the cumulative prefix.
Later levels add detail that is absent from the 8-channel prefix
(Figure~\ref{fig:staged_generation};
Appendix~\ref{app:qual_generations}).
Successive-prefix differences are shown in
Appendix~\ref{app:qual_differences}.

\paragraph{DiT initialization.}
We initialize the rows of the input projection associated with
\(b_0\) using the standard SiT initialization \citep{ma2024sit}
and the rows associated with \(b_1,\ldots,b_{L-1}\) to zero.
The model therefore begins as an \(L_0\)-only generator and
gradually learns to incorporate refinement information, without
modifying the Transformer trunk.

\begin{figure}[t]
\centering
\includegraphics[width=\linewidth]
{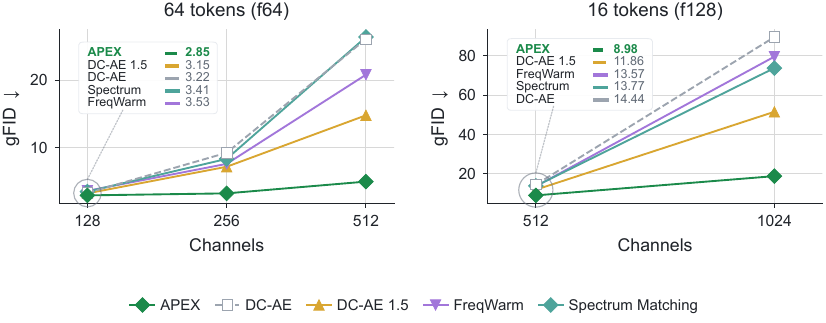}
\caption{\textbf{APEX degrades far more slowly as latent width increases.}
Left: 64-token f64 latents; right: 16-token f128 latents.
At f64, APEX changes from \(2.85 \rightarrow 3.15 \rightarrow 4.90\) gFID
as channels increase \(128 \rightarrow 256 \rightarrow 512\),
while DC-AE degrades from \(3.22 \rightarrow 9.19 \rightarrow 26.10\).
At f128, increasing channels from \(512 \rightarrow 1024\) changes
APEX from \(8.98 \rightarrow 18.77\), compared with
DC-AE \(14.44 \rightarrow 89.47\) and DC-AE 1.5 \(11.86 \rightarrow 51.47\).
The boxed values summarize performance at f64c128 and f128c512.
All runs use matched denoiser-training FLOPs and total NFE.}
\label{fig:channel_scaling}
\end{figure}

\section{Experiments}
\label{sec:experiments}
\label{sec:experimental_setup}

We target the operating point where each token must absorb substantial
instance-specific detail: high native resolution under aggressive spatial
compression. Following DC-AE \citep{chen2025dcae}, we evaluate on
ImageNet-512 \citep{deng2009imagenet} at \(64\times\) and \(128\times\) spatial compression.
We first compare
reconstruction and generation across f64 and f128 latents with
increasing channel capacity, then test whether high compression remains
advantageous when both training and inference are compute-constrained.
We next perform an extensive
component study showing that the gain emerges from the complete
representation--diffusion co-design: neither the hierarchical latent
nor asynchronous denoising reaches the same quality on its own.
Finally, we study the schedule choices used by the main
models.

\paragraph{Setup.}
We denote an autoencoder with \(f\times\) spatial compression and
\(C\) channels by f\(f\)c\(C\). At \(512^2\), f32, f64, and f128
produce \(16\times16\), \(8\times8\), and \(4\times4\) latent grids,
corresponding to 256, 64, and 16 Transformer tokens.
APEX uses c\(8{-}32{-}128\) at f64 and
c\(32{-}128{-}256{-}512\) at f128; wider variants append c256 and
c512 at f64 and c1024 at f128. Appendix
Table~\ref{tab:app_hierarchies} gives the exact groups and target
resolutions.
Unless stated otherwise, we use SiT-XL/1 \citep{ma2024sit}, with SiT-B/1 for component and design ablations. We report rFID on the ImageNet validation set and gFID from 50K class-conditional samples using the ADM protocol \citep{dhariwal2021adm}. Generation uses Euler sampling and CFG \citep{ho2022cfg} with scale 1.5 unless otherwise specified; Appendix~\ref{app:numerical} additionally reports unguided gFID, sFID, and Inception Score. Within each comparison, we match cumulative denoiser-training FLOPs and, unless varied, total NFEs; methods sharing a latent shape also use the same denoiser, optimizer, batch, and evaluation protocol. Full configurations and compute accounting are in Appendix~\ref{app:training_details}.
\paragraph{Baselines.}
DC-AE uses the unstructured reconstruction latent of
\citet{chen2025dcae}. DC-AE 1.5 trains cumulative channel prefixes and
augments diffusion with randomly sampled prefixes under a shared
timestep \citep{chen2025dcae15}. FreqWarm changes the frequency
curriculum of the diffusion model without modifying the autoencoder
\citep{lai2025freqwarm}. Spectrum Matching regularizes the latent and
reconstruction spectra \citep{ning2026spectrum}. We implement all
baselines in the same DC-AE and SiT training framework.

\subsection{Generation Across Channel Counts}
\label{sec:channel_scaling}

\begin{wraptable}{r}{0.48\linewidth}
\vspace{-6pt}
\centering
\small
\setlength{\tabcolsep}{2pt}
\caption{\textbf{Matched full-latent reconstruction quality.}
Every method at a given latent geometry is trained to the same rFID shown below.}
\label{tab:channel_rfid}
\begin{tabular*}{\linewidth}{@{\extracolsep{\fill}}lccc@{}}
\toprule
Latent & \shortstack{Matched\\rFID \(\downarrow\)} & PSNR \(\uparrow\) & Reference \\
\midrule
f64c128   & 0.22 & 26.15 & Official DC-AE \\
f64c256   & 0.18 & 26.8 & Trained by us \\
f64c512   & 0.13 & 27.30 & Trained by us \\
f128c512  & 0.23 & 25.73 & Official DC-AE \\
f128c1024 & 0.20 & 26.1 & Trained by us \\
\bottomrule
\end{tabular*}
\vspace{-5pt}
\end{wraptable}
Following DC-AE 1.5 \citep{chen2025dcae15}, we find that imposing a
channel hierarchy has little effect on complete-latent reconstruction,
so we match all methods at each latent geometry to the same rFID.
We use the official DC-AE f64c128 and f128c512 models as reference
points and train flat DC-AE models at the additional widths using the
same architecture and recipe (Table~\ref{tab:channel_rfid}).
We evaluate c128, c256, and c512 at f64 and c512 and c1024 at f128,
matching cumulative denoiser-training FLOPs and total NFE across
methods.

Figure~\ref{fig:channel_scaling} shows that APEX degrades far more slowly as the latent widens. At f64, APEX gFID changes
only from \(2.85\) at c128 to \(3.15\) at c256 and \(4.9\) at c512,
while DC-AE rises from \(3.22\) to \(9.19\) and \(26.1\). At f128,
APEX reaches \(8.98\) and \(18.77\) at c512 and c1024, versus
\(14.44/89.47\) for DC-AE and \(11.86/51.47\) for DC-AE 1.5.
At f128c1024, Spectrum Matching and FreqWarm reach \(73.71\) and
\(79.51\), respectively.

Unguided APEX results across channel counts are in
Appendix~\ref{app:channel_metrics}.

\subsection{APEX Makes High Compression Advantageous Under Constrained Compute}
\label{sec:joint_compute}

\begin{figure}[t]
\centering
\includegraphics[width=\linewidth]
{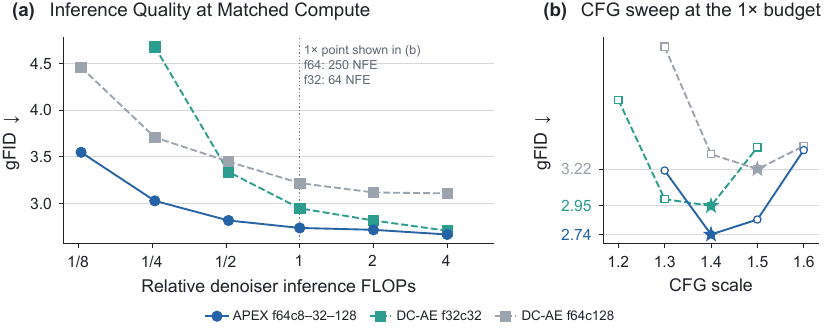}
\caption{\textbf{APEX makes high compression advantageous under constrained compute.}
\textbf{(a)} Guided gFID versus relative denoiser inference FLOPs, where
\(1\times\) matches f64 at 250 NFE with f32 at 64 NFE.
\textbf{(b)} CFG sweep at the \(1\times\) budget. Stars mark the selected
guidance scale for each model.}
\label{fig:inference_compute}
\end{figure}

\begin{wraptable}[16]{r}{0.53\linewidth}
\vspace{-14pt}
\centering
\captionsetup{justification=raggedright,singlelinecheck=false,font=footnotesize,skip=1pt}
\caption{\textbf{Building APEX from DC-AE.}
The first block successively introduces APEX components; the
second removes components from the complete method. All runs use
f64c128 and SiT-B/1 at matched compute, and
\(\Delta\)gFID is measured relative to APEX.}
\label{tab:component_ablation}
\footnotesize
\setlength{\tabcolsep}{2.5pt}
\renewcommand{\arraystretch}{1.02}
\begin{tabularx}{\linewidth}{@{}>{\raggedright\arraybackslash}Xrr@{}}
\toprule
Configuration
& gFID \(\downarrow\)
& \(\Delta\)gFID \\
\midrule
\multicolumn{3}{@{}l}{\emph{Build APEX from DC-AE}} \\[1pt]
DC-AE
& 13.27
& \dgfid{5.15} \\
\hspace{0.8em}\(\boldsymbol{+}\) APEX levels
& 20.37
& \dgfid{12.25} \\
\hspace{0.8em}\(\boldsymbol{+}\)
cumulative-prefix diffusion training
& 17.83
& \dgfid{9.71} \\
\hspace{0.8em}\(\boldsymbol{+}\)
\textbf{APEX diffusion}
& \textbf{8.12}
& \textbf{0.00} \\
\addlinespace[2pt]
\multicolumn{3}{@{}l}{\emph{Remove components from APEX}} \\[1pt]
\hspace{0.8em}\(\boldsymbol{-}\)
matched-resolution targets
& 8.77
& \dgfid{0.65} \\
\hspace{0.8em}\(\boldsymbol{-}\)
APEX levels and clock sampler
\newline
\hspace*{2.05em}\emph{(DCAE1.5 + naive async schedule)}
& 52.89
& \dgfid{44.77} \\
\bottomrule
\end{tabularx}
\end{wraptable}
APEX turns the lower per-step cost of extreme compression into a genuine
quality advantage. We compare APEX f64c\(8{-}32{-}128\) with DC-AE
f64c128 and f32c32 baselines using the same denoiser architecture and
matched cumulative denoiser-training FLOPs. The f64 models are trained
for 500K steps with batch size 2048, while f32c32 is trained for 250K
steps with batch size 1024. One f32 denoiser evaluation costs
approximately \(4\times\) one f64 evaluation, so a matched inference
budget affords the f64 models approximately four times as many NFEs.
For each model, we select the best evaluated CFG value independently at each inference budget. Figure~\ref{fig:inference_compute}(b)
shows the full sweep at the \(1\times\) operating point, where APEX and
f32c32 select CFG \(1.4\), while flat f64c128 selects \(1.5\).
\WFclear

Figure~\ref{fig:inference_compute}(a) shows that compression alone does
not guarantee a better quality--compute trade-off. The flat f64
baseline is preferable only at the smallest sampling budgets; by
\(1/2\times\) relative compute, f32c32 is already better despite
receiving one quarter as many evaluations. APEX removes this crossover:
it outperforms f32c32 at every matched guided inference budget, with the
margin increasing as compute becomes more constrained. At the
\(1\times\) budget, APEX reaches a gFID of \(2.74\) with 250 NFEs,
compared with \(2.95\) for f32c32 with 64 NFEs.
At 1000 NFEs (\(4\times\)), APEX reaches \(2.67\).
APEX therefore makes high compression
advantageous across the full evaluated compute-constrained regime.
\raggedbottom

\subsection{The Hierarchy and Denoising Process Must Work Together}
\label{sec:component_analysis}
\label{sec:ablation_resolution}

\begin{wrapfigure}[16]{r}{0.48\linewidth}
\vspace{-16pt}
\centering
\includegraphics[width=\linewidth]{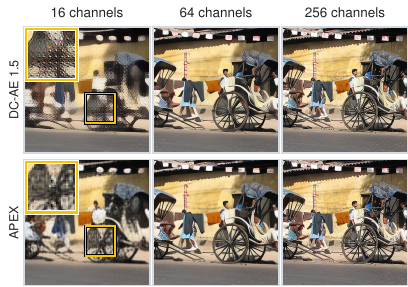}
\captionsetup{justification=raggedright,singlelinecheck=false,font=footnotesize}
\caption{\textbf{Prefix reconstructions.} DC-AE 1.5 (top) vs APEX (bottom)
prefixes. Insets show the same crop at 16 channels. High-frequency detail leaks into DC-AE 1.5's early levels.}
\label{fig:prefix_reconstructions}
\end{wrapfigure}
The complete APEX design delivers a clear performance jump over
every partial construction. We extensively ablate the representation, autoencoder
targets, channel partition, diffusion schedule, loss aggregation, and
training sampler at f64c128 using SiT-B/1. All variants use the same batch, evaluation setting, and cumulative
denoiser-training FLOPs. We study representative controls
obtained by using our hierarchy without asynchronous generation, and
by applying asynchronous generation without the matching hierarchy or
sampling regime.

Here, APEX levels denote f64c8-32-128 trained with matched-resolution
targets. APEX diffusion combines staggered local times, level-balanced
losses, and the APEX clock sampler. Table~\ref{tab:component_ablation}
first builds APEX from DC-AE, then removes key components from the
complete method.
Appendix~\ref{app:ablation_metrics} reports sFID and IS for the same
variants.

The cumulative path shows that the hierarchy alone is insufficient.
Adding APEX levels to ordinary diffusion worsens gFID, and
cumulative-prefix training recovers only part of this loss. The large
improvement appears when the matching diffusion process is introduced,
reducing gFID from 17.83 to 8.12. The subtractive controls tell the same story. Removing
matched-resolution targets raises gFID to 8.77, while naively replacing the
APEX levels and clock sampler with the DC-AE 1.5 construction of many small uniform-width levels fails
severely at 52.89. Figure~\ref{fig:prefix_reconstructions} shows the
corresponding representational difference: APEX prefixes are coherent
and usable at lower resolutions, whereas DC-AE 1.5 prefixes mix coarse
structure with isolated high-frequency detail.
\WFclear

\subsection{Design Choices}
\label{sec:ablations}
\label{sec:critical_design}

Two choices are critical in practice: the stagger gap and the allocation
of training probability along the global clock. Unless stated otherwise, these experiments use
f64c128 and SiT-B/1. Only the denoiser size and the explicitly ablated variable change.
Appendix~\ref{app:ablation_metrics} additionally reports unguided gFID
for the gap and allocation sweeps, together with sFID and IS for all
three analyses.

\begin{figure}[t]
\vspace{-8pt}
\centering
\includegraphics[width=\linewidth]{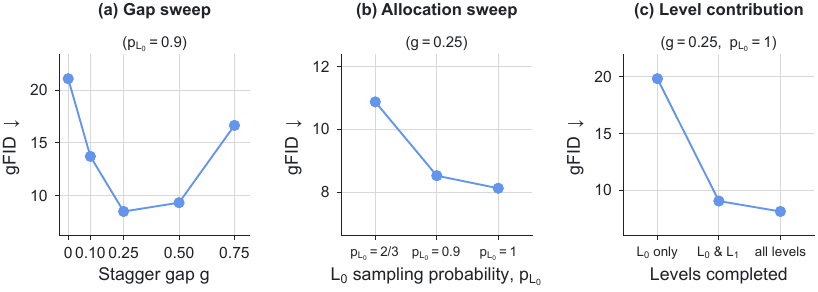}
\caption{\textbf{Gap, allocation, and level contribution.}
\textbf{(a)} We vary the stagger gap while using \(p_{L_0}=0.9\). At \(g=0\), the tail interval is absent and the process reduces to synchronous diffusion.
\textbf{(b)} Holding \(g=0.25\), we vary \(p_{L_0}\), the training probability assigned to the interval where \(L_0\) is denoising. Uniform clock sampling corresponds to \(p_{L_0}=2/3\).
\textbf{(c)} For the \(g=0.25,\ p_{L_0}=1\) checkpoint, we stop inference when \(L_0\), \(L_1\), or \(L_2\) completes and redistribute the same 250 NFEs over the retained clock interval. Generation improves at every endpoint, showing that all three levels contribute.}
\label{fig:schedule_ablations}
\end{figure}

\paragraph{Stagger gap.}
Increasing \(g\) gives later levels cleaner structural context, but under
our equal-gap schedule it also lengthens the global trajectory and
reduces overlap, making a fixed NFE budget less efficient.
Figure~\ref{fig:schedule_ablations}(a) shows that this trade-off is
optimized near \(g=0.25\).

\paragraph{Training allocation.}
\(L_0\) carries the unconditional generation burden, whereas later
levels solve easier conditional refinements. We therefore place mass
\(p_{L_0}\) on the interval where \(L_0\) is active and distribute the
remainder uniformly over the rest of the clock
(Figure~\ref{fig:clock_density}).
Figure~\ref{fig:schedule_ablations}(b) shows that increasingly
structure-heavy allocation improves gFID, with \(p_{L_0}=1\) performing
best even though later portions of the clock receive no direct
supervision.
For the main models, we use \(p_{L_0}=0.9\) rather than \(1\).
The \(p_{L_0}=1\) result above uses only three levels; in deeper
hierarchies, later levels begin at or after \(\tau=1\) and would receive
little or no direct supervision. Using \(0.9\) preserves tail coverage
while still concentrating training on structure generation.

To quantify the contribution of each level, we evaluate the
\(p_{L_0}=1\) model with the same 250-NFE budget redistributed over
successively longer prefixes of its three-level clock. Stopping after
the first one, two, or three levels reach \(t=0\), and decoding the
corresponding cumulative latent, gives gFID \(19.83\), \(9.04\), and
\(8.12\), respectively. These results confirm that the model uses the
full hierarchy: later levels require far less direct supervision than
\(L_0\), yet remain essential to final generation quality. In
particular, \(L_2\) still contributes despite receiving no direct
supervision over the latter half of its trajectory and despite the
modest SiT-B/1 denoiser used here.

\section{Conclusion and Limitations}
\label{sec:limitations}

We introduced APEX, which pairs a resolution-aware hierarchy of channel
prefixes with asynchronous denoising across channels. At \(64\times\)
and \(128\times\) spatial compression, APEX substantially shifts the
reconstruction--generation frontier as channels are added. With matched
training compute, it also outperforms a less-compressed 256-token
baseline across all evaluated guided inference budgets, making extreme
compression competitive under constrained compute.

APEX still requires choosing channel boundaries, per-level target
resolutions, the stagger gap, and the global-clock distribution.
Section~\ref{sec:ablations} provides practical guidance, but the optimal
choices may vary across settings. Learning them jointly with the
autoencoder and denoiser is a promising direction.

\nopagebreak
\subsection*{Reproducibility}
The appendices provide complete autoencoder and diffusion training
details, latent configurations, sampling and evaluation protocols,
FLOP accounting, training and inference pseudocode, and figure-specific
reproduction instructions.
Appendix~\ref{app:figure_reproduction} records the protocol and data
required to reproduce each main-text figure. Code is available at
\url{https://daviddinkevich.github.io/APEX}.

\subsection*{Acknowledgements}

This work was supported in parts by the Israel Science
Foundation (grant no.~2203/24).

\bibliography{apex}
\bibliographystyle{apex}

\clearpage
\appendix
\raggedbottom
\captionsetup{justification=raggedright,singlelinecheck=false,font=small}
\begingroup
\newcommand{\apexGalleryTile}[3]{%
  \parbox[c][#2][c]{#2}{\centering\apexSquare{#1}{#2}{#3}}}
\newcommand{\apexSampleTag}[2]{%
  \parbox[c][#2][c]{8pt}{\centering%
  \rotatebox[origin=c]{90}{\fontsize{8}{9}\selectfont Sample #1}}}
\newcommand{\apexGalleryLabel}[1]{{\fontsize{8}{10}\selectfont #1}}
\section{Additional Visualizations}
\label{app:qualitative}
\subsection{Generated prefixes}
\label{app:qual_generations}
Each row is one generated sample. Columns show the image decoded when that level first reaches \(t=0\).

\noindent
\includegraphics[width=\linewidth]{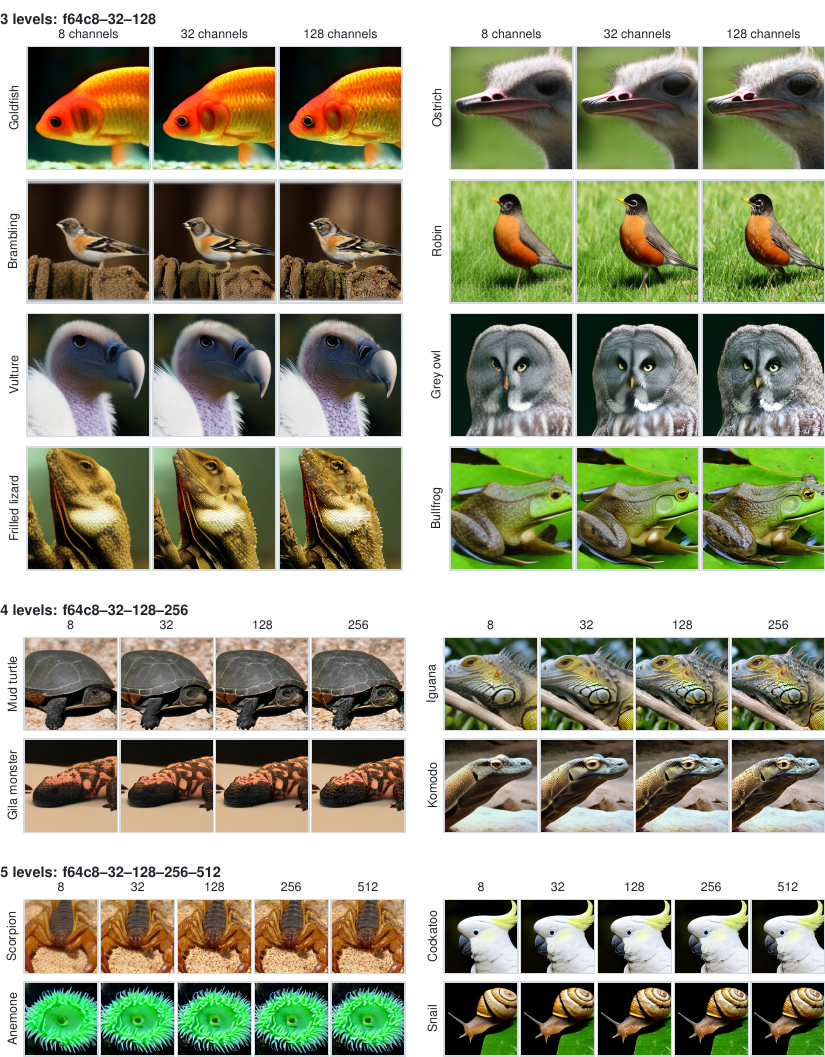}
\captionof{figure}{\textbf{Generated prefixes.} Each row is one sample from the staggered generator. We decode the completed prefix when each successive level reaches \(t=0\).}
\label{fig:app_generations}

\clearpage
\subsection{Refinement differences}
\label{app:qual_differences}
For the two generated examples in Figure~\ref{fig:staged_generation}, we show
successive prefix decodes and their differences. Here $x_0$, $x_1$, and $x_2$
are the images decoded from 8, 32, and 128 channels.

\noindent\begin{minipage}{\linewidth}
\setlength{\parskip}{0pt}
\noindent{\small\textbf{Sample 1}}\par\vspace{5pt}
\setlength{\tabcolsep}{0pt}\begin{tabular*}{\linewidth}{@{}c@{\extracolsep{\fill}}cccc@{}}
{\small 8 channels} & {\small 32 channels} & {\small 128 channels} & {\small $x_1-x_0$} & {\small $x_2-x_1$}\\[5pt]
\apexGalleryTile{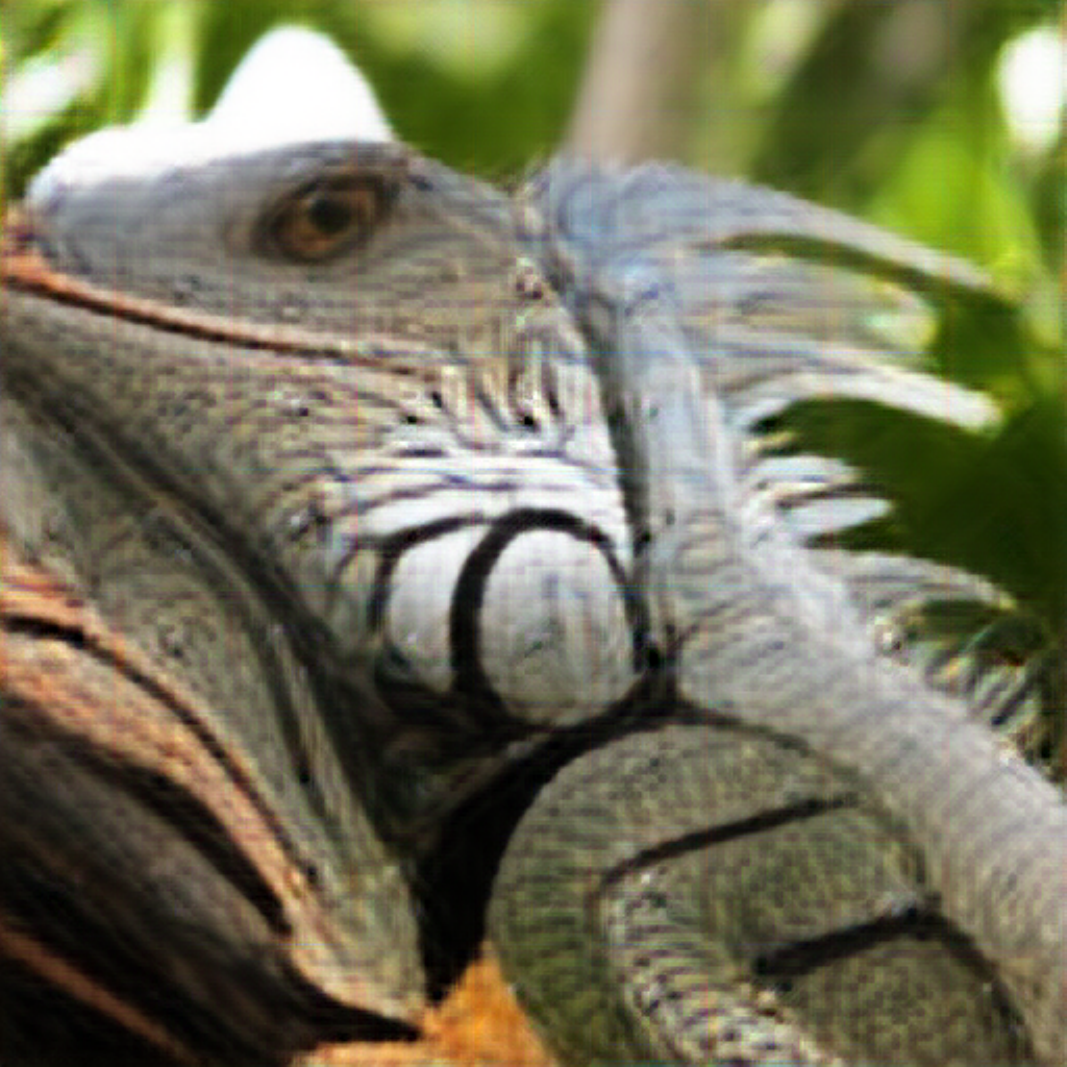}{1.02in}{8 channels} & \apexGalleryTile{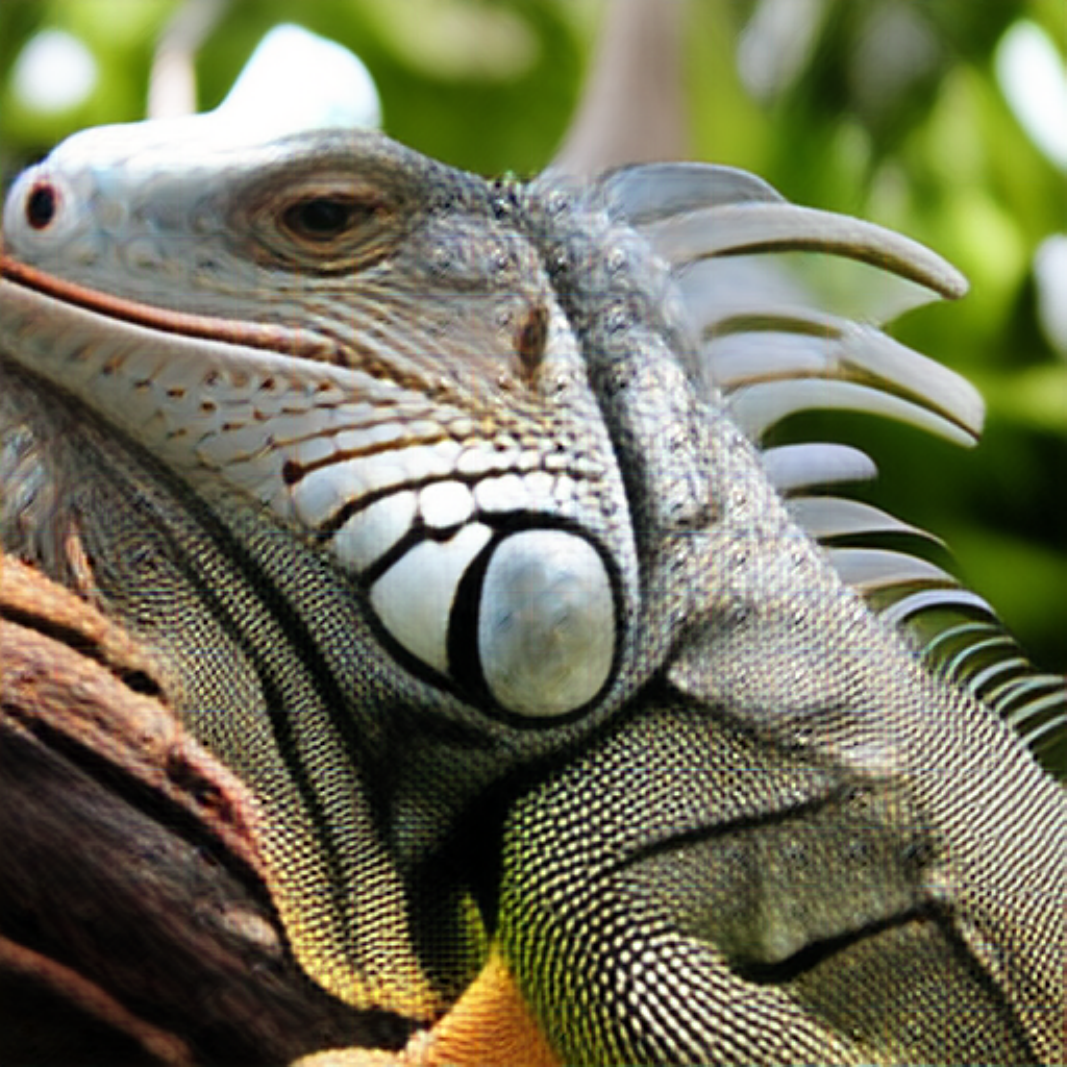}{1.02in}{32 channels} & \apexGalleryTile{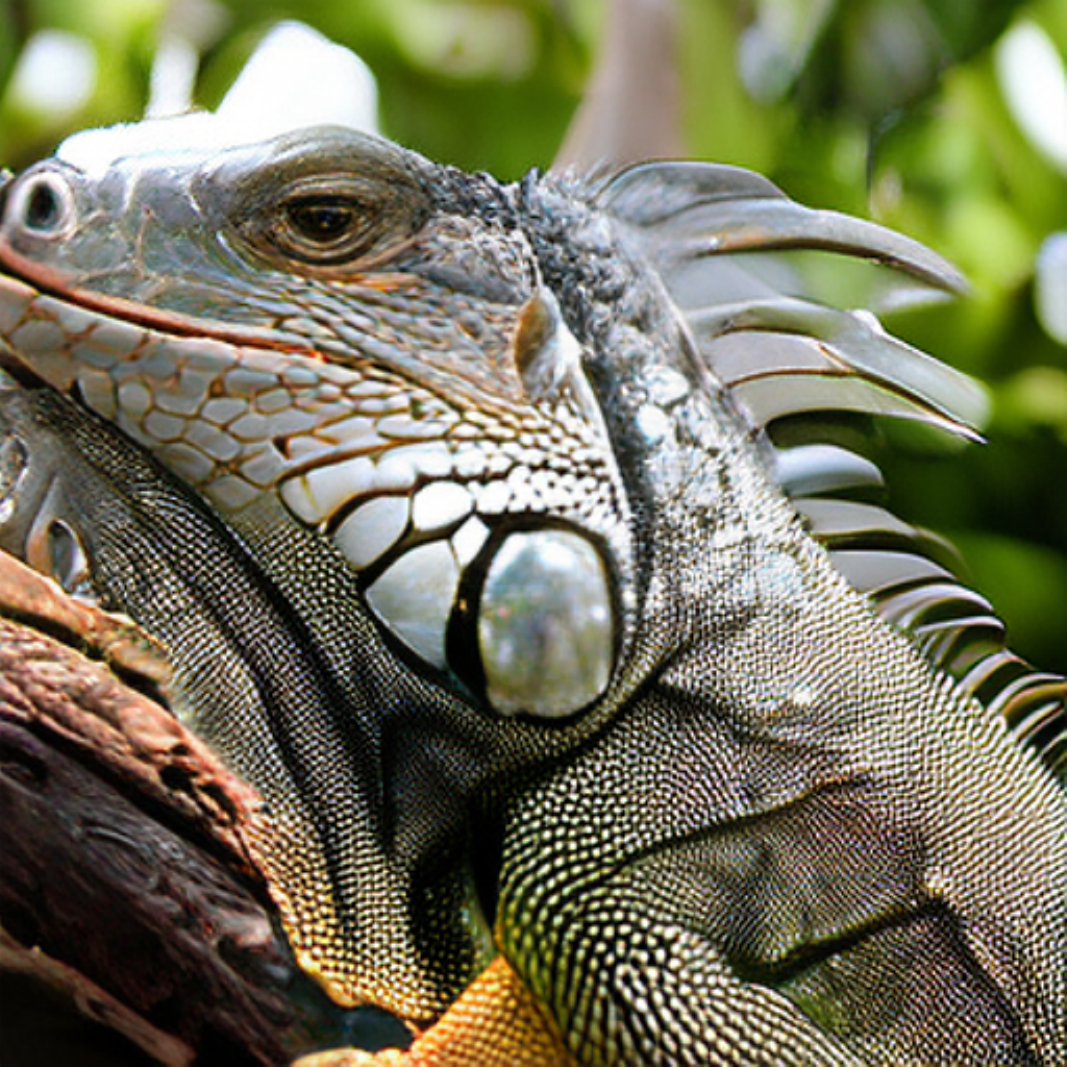}{1.02in}{128 channels} & \apexGalleryTile{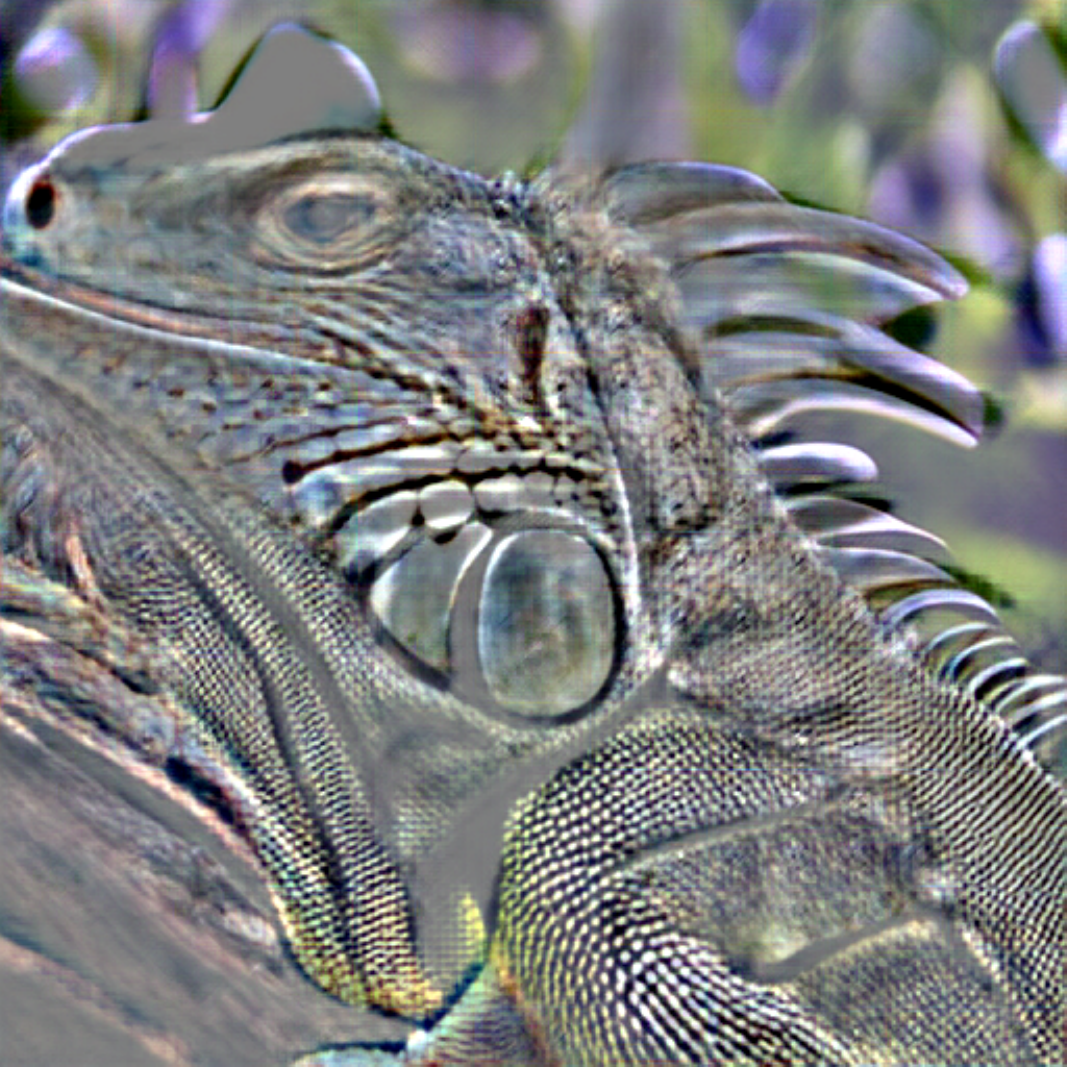}{1.02in}{difference} & \apexGalleryTile{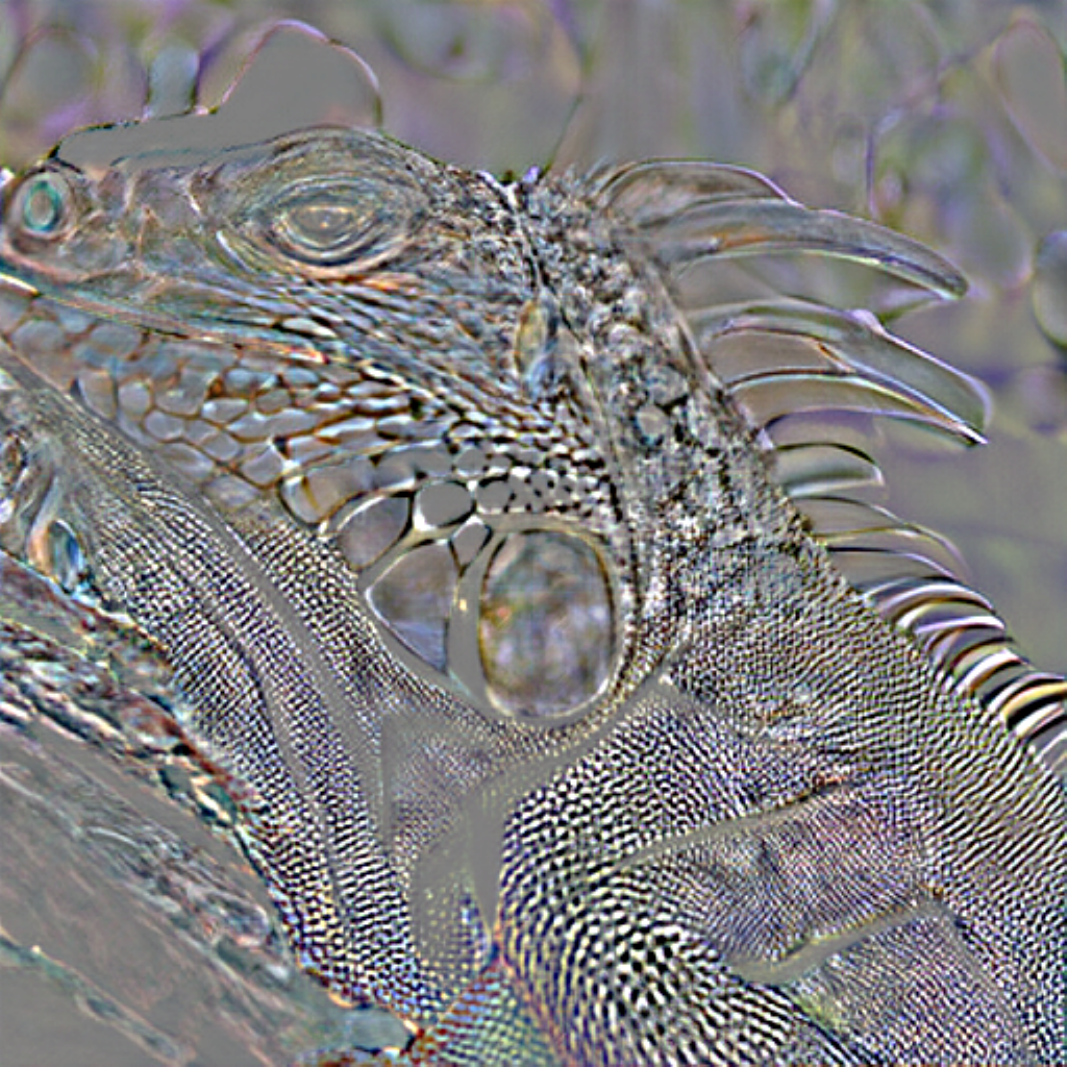}{1.02in}{difference}\\
\end{tabular*}
\par\vspace{13pt}
\noindent{\small\textbf{Sample 2}}\par\vspace{5pt}
\setlength{\tabcolsep}{0pt}\begin{tabular*}{\linewidth}{@{}c@{\extracolsep{\fill}}cccc@{}}
{\small 8 channels} & {\small 32 channels} & {\small 128 channels} & {\small $x_1-x_0$} & {\small $x_2-x_1$}\\[5pt]
\apexGalleryTile{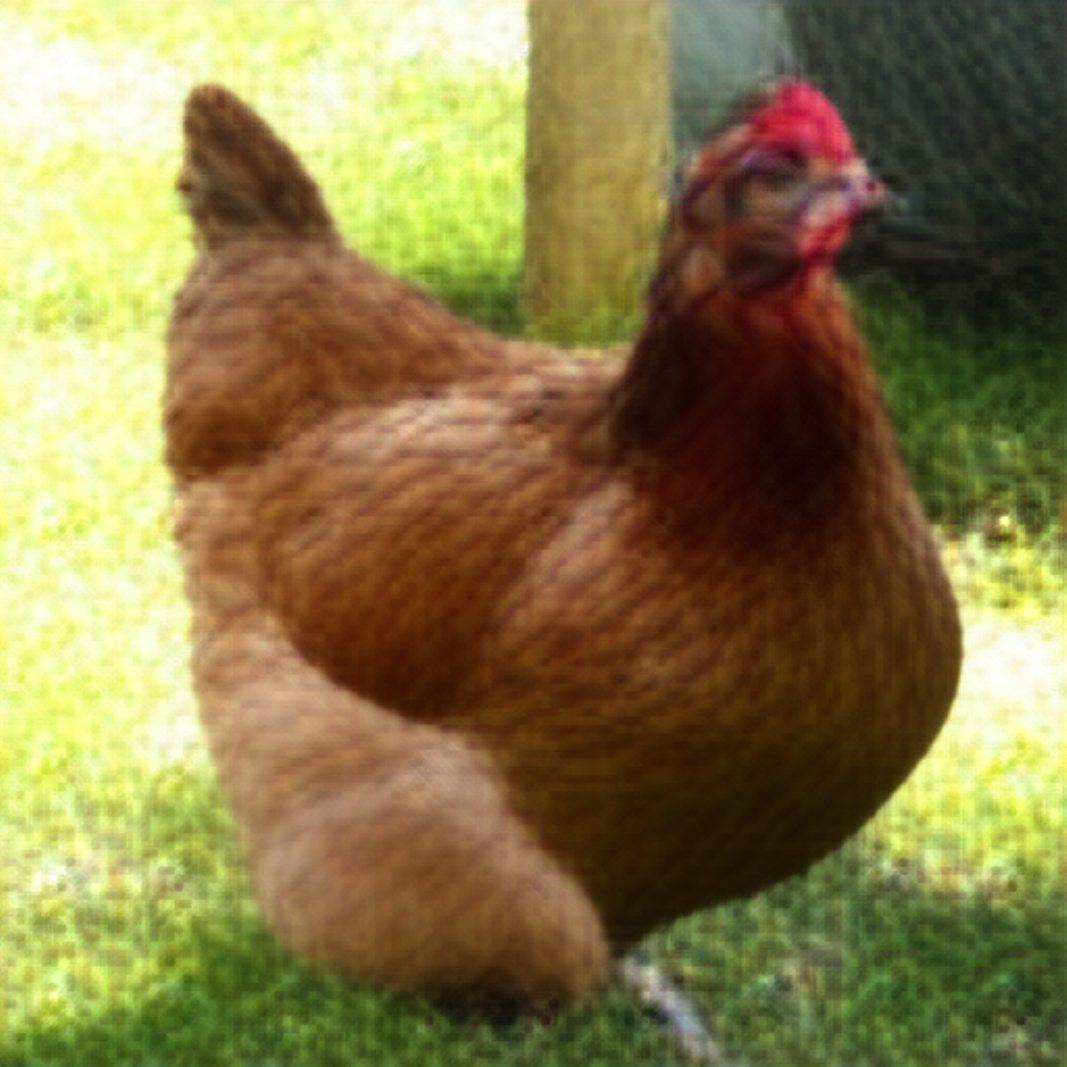}{1.02in}{8 channels} & \apexGalleryTile{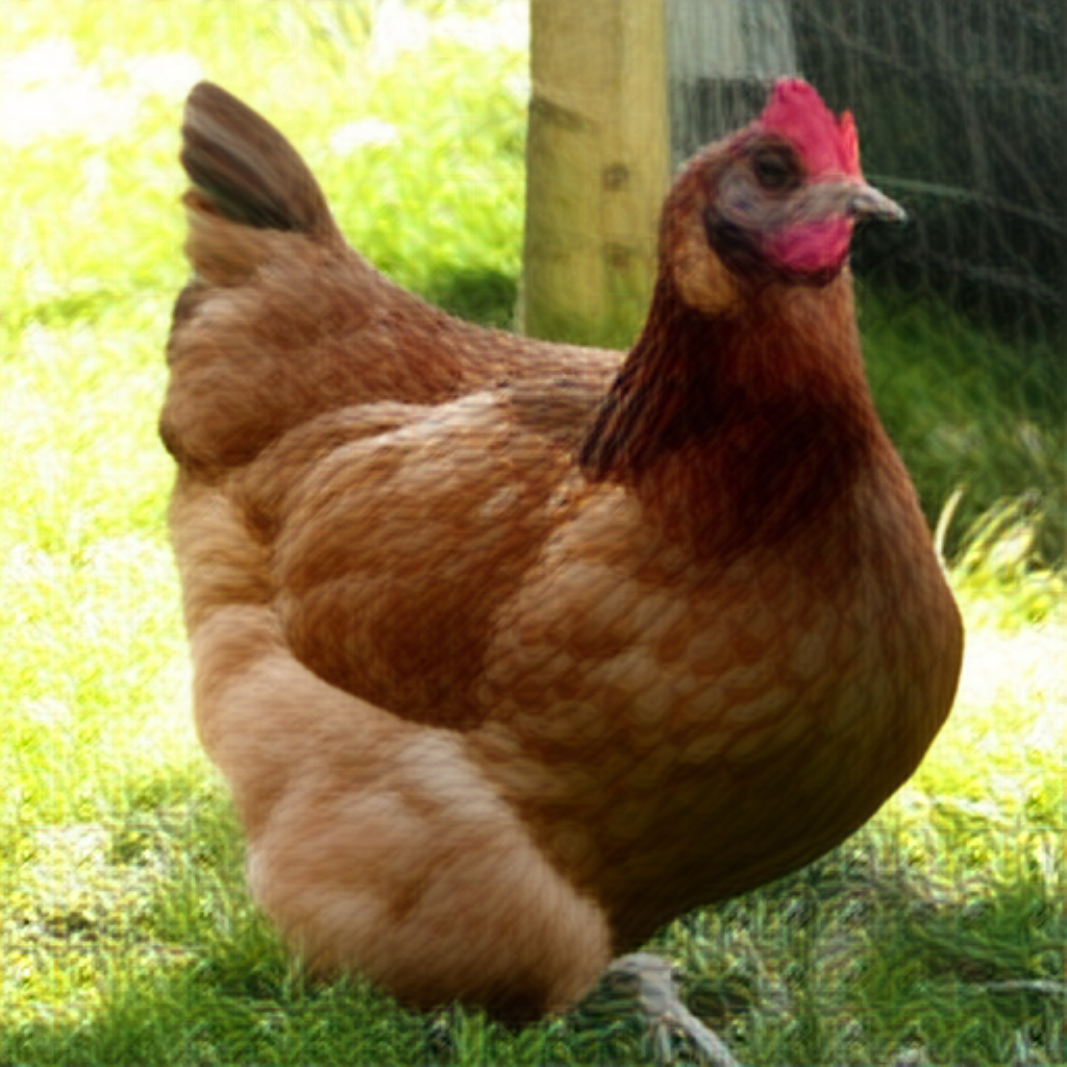}{1.02in}{32 channels} & \apexGalleryTile{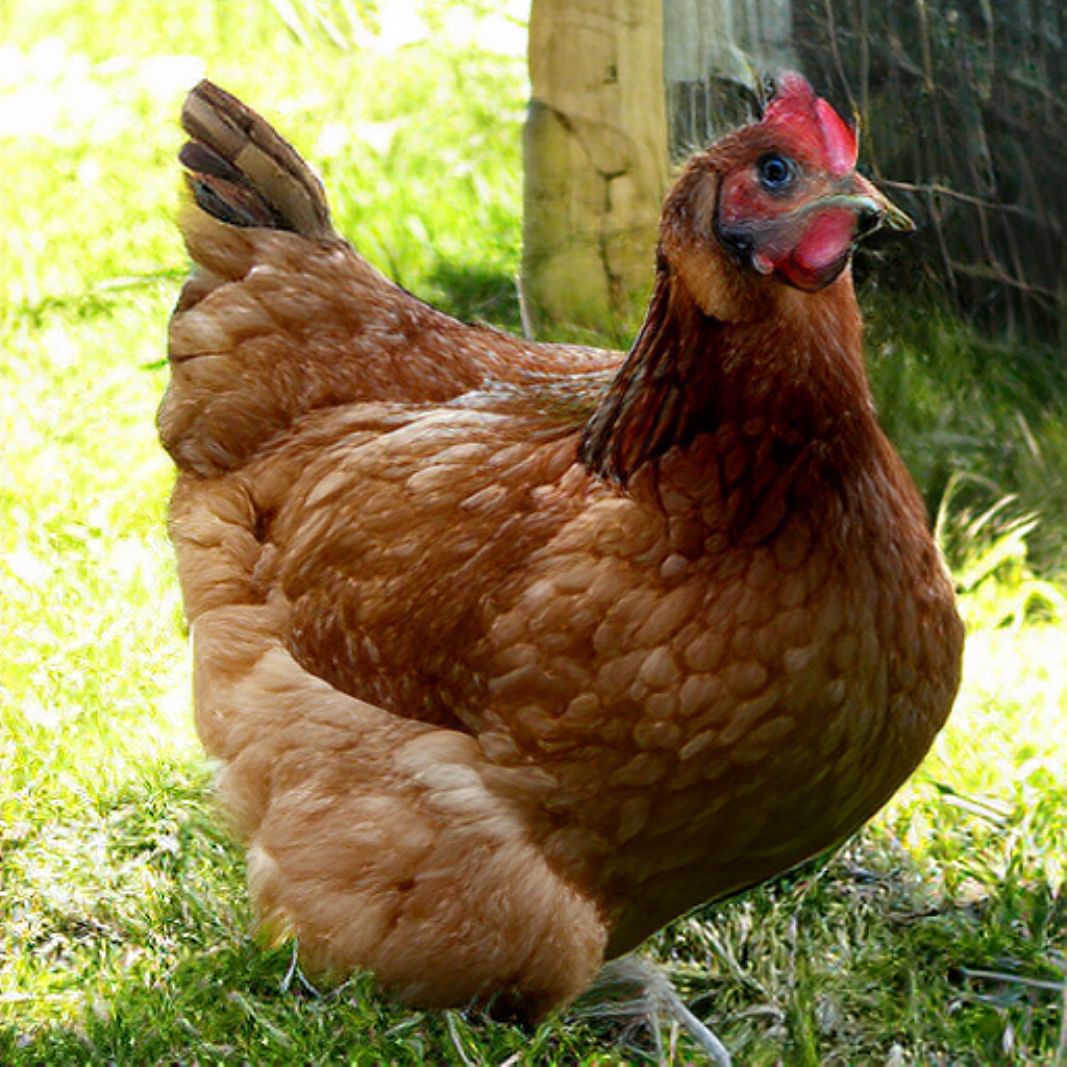}{1.02in}{128 channels} & \apexGalleryTile{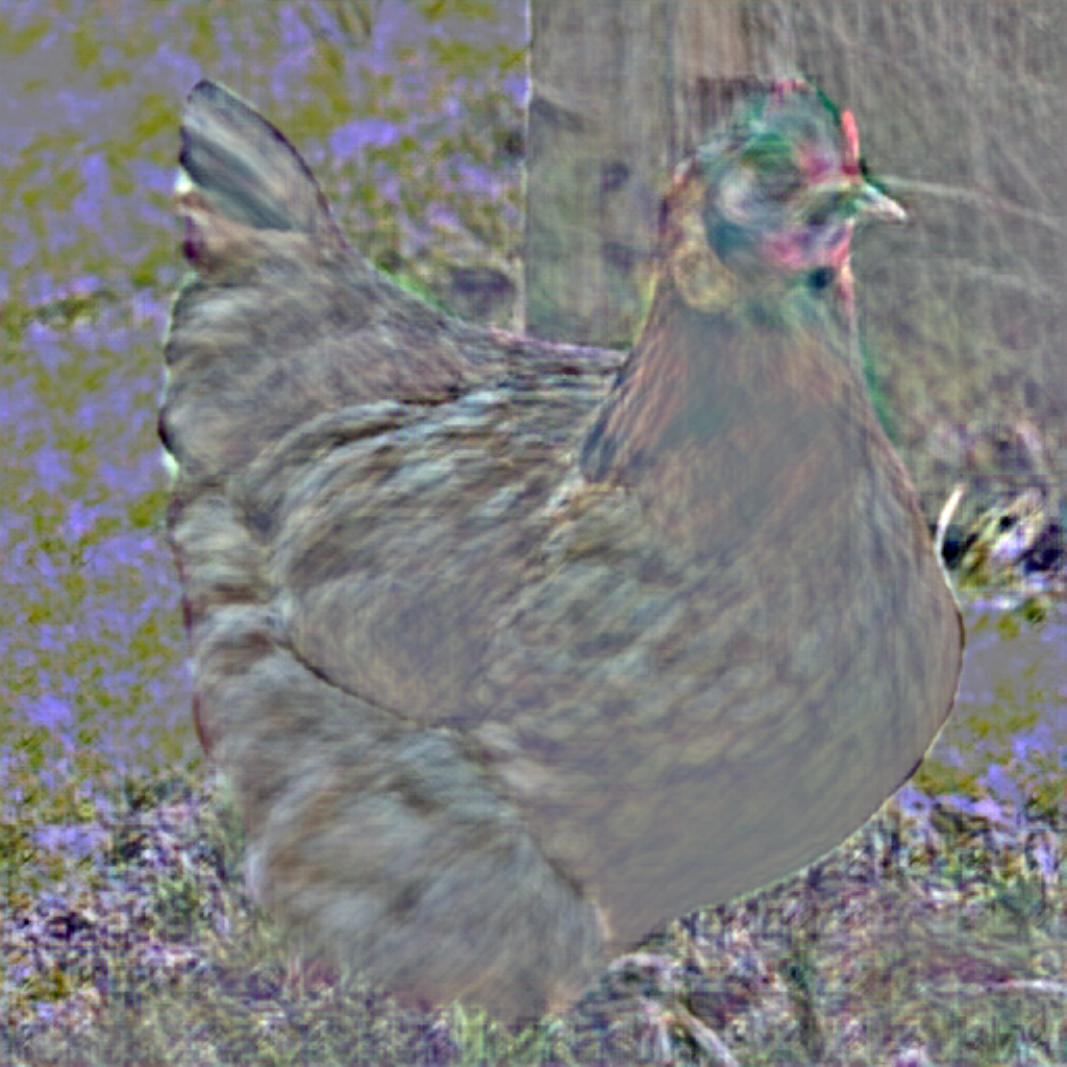}{1.02in}{difference} & \apexGalleryTile{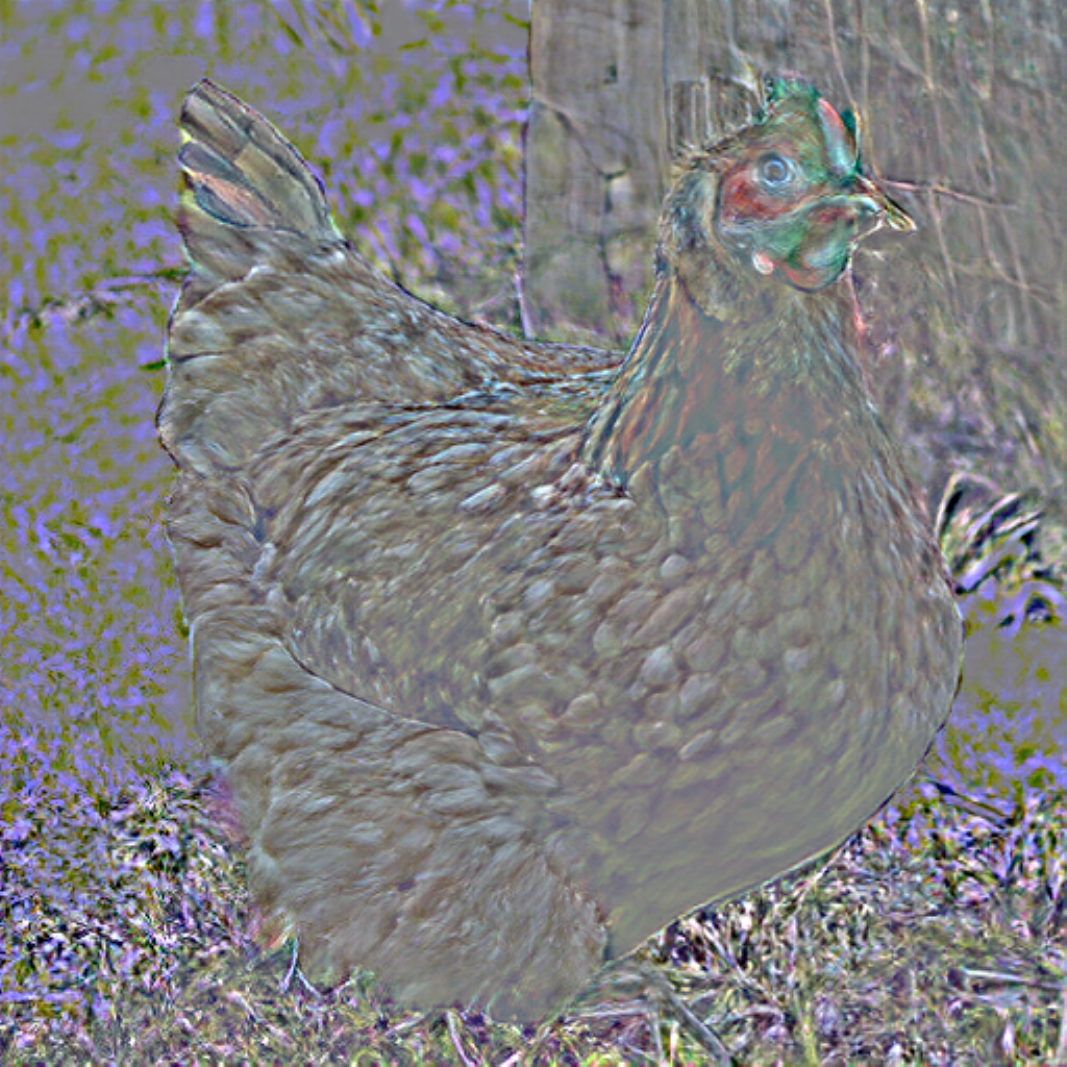}{1.02in}{difference}\\
\end{tabular*}

\par\vspace{7pt}
\captionof{figure}{Successive generated prefixes and the changes introduced by
later levels. Both samples use the same difference scale, with zero shown as gray.}
\label{fig:app_differences}
\end{minipage}

Differences are computed per channel on the full-resolution decodes, and mapped
to gray-centered images with a shared display scale of $\pm0.5$ in the $[0,1]$
RGB range, which clips $1.3\%$ of pixels.

\endgroup

\section{Extended Experimental Details}
\label{app:training_details}

\subsection{Compute-matched training}
Within each experiment, we match \emph{cumulative denoiser-training FLOPs}
across all compared methods. This applies when changing the spatial compression,
channel count, or denoiser size. We adjust the number of training steps for the
model cost and global batch size; the comparison is not based on equal numbers
of updates or images seen.

Table~\ref{tab:app_compute} gives the schedules for the main SiT-XL/1
comparisons. Figure~\ref{fig:channel_scaling} uses the f64 and f128 settings,
and Figure~\ref{fig:inference_compute} uses the f32 and f64 settings.
All three schedules correspond to approximately $1.8\times10^{20}$
denoiser-training FLOPs.

\begin{table}[ht]
\centering\small
\caption{Main diffusion-training schedules for ImageNet-512. Batch sizes are global.}
\label{tab:app_compute}
\begin{tabular*}{\linewidth}{@{}l@{\extracolsep{\fill}}rrrr@{}}
\toprule
Latent & Tokens & Batch size & Training steps & Training FLOPs\\
\midrule
f32c32 & 256 & 1,024 & 250,000 & $\approx1.8\times10^{20}$\\
f64, all widths & 64 & 2,048 & 500,000 & $\approx1.8\times10^{20}$\\
f128, all widths & 16 & 4,096 & 1,000,000 & $\approx1.8\times10^{20}$\\
\bottomrule
\end{tabular*}
\end{table}

\paragraph{FLOP calculation.}
Following DiT \citep{peebles2023dit}, we compute the training cost arithmetically as
\begin{equation}
F_{\mathrm{train}}\simeq 3F_{\mathrm{forward}}BS,
\label{eq:app_flops}
\end{equation}
where $F_{\mathrm{forward}}$ is the cost of one denoiser forward pass on one
image, $B$ is the global batch size, and $S$ is the number of updates.
A multiply-add counts as two FLOPs. The factor of three approximates the
forward and backward passes. At these sequence lengths, the dominant cost is
approximately linear in the number of tokens. Autoencoder training is separate
from this budget.

\subsection{Models and optimization}
We train on the ImageNet-1K training split at $512\times512$ \citep{deng2009imagenet}. Main experiments
use class-conditional SiT-XL/1; component and schedule ablations use SiT-B/1.
Patch size 1 assigns one token to each latent spatial location.
We follow the DC-AE diffusion-training setup and the underlying SiT optimization
recipe \citep{chen2025dcae,ma2024sit}: AdamW \citep{loshchilov2019adamw} with a constant learning rate of
$10^{-4}$ and an exponential moving average with decay $0.9999$.
Batch sizes and training lengths are given in Table~\ref{tab:app_compute}.

\paragraph{Baselines.}
All methods use the same DC-AE and SiT training framework. DC-AE uses an
unstructured latent \citep{chen2025dcae}; DC-AE 1.5 uses cumulative prefixes
and shared-timestep prefix training \citep{chen2025dcae15}. FreqWarm applies
its frequency curriculum to diffusion training \citep{lai2025freqwarm}, and
Spectrum Matching regularizes the latent and reconstruction spectra
\citep{ning2026spectrum}. Within a comparison, the backbone, optimizer,
training compute, and evaluation protocol are shared.

\subsection{Sampling and evaluation}
We evaluate generation using 50,000 class-conditional samples and the ADM
protocol \citep{dhariwal2021adm}, reporting gFID \citep{heusel2017fid}, sFID \citep{nash2021sfid}, and Inception Score
(IS) \citep{salimans2016improved}. We use classifier-free guidance with scale 1.5 unless another scale is
specified \citep{ho2022cfg}. Unguided sampling retains the class label but does
not apply guidance (CFG 1.0). We report unguided gFID for the channel-scaling
comparisons wherever the guided checkpoint is identified, and for the SiT-B
gap and clock-allocation ablations, using the same trained model and number of
sampling steps as the guided evaluation. Reconstruction rFID and PSNR are
evaluated on the ImageNet validation split using the complete latent, following
DC-AE \citep{chen2025dcae}.

For $N$ sampling steps, denoiser inference cost is $NF_{\mathrm{forward}}$
without guidance and $2NF_{\mathrm{forward}}$ with guidance, excluding the
autoencoder decoder. One f32 evaluation costs approximately four times an f64
evaluation. The $1\times$ budget in Figure~\ref{fig:inference_compute} therefore
pairs 64 f32 evaluations with 250 f64 evaluations. APEX sampling follows the
Euler update in Section~\ref{app:sampling}; sampling budgets are given with
the individual comparisons.

\subsection{Comparison to f8c4 SD-VAE}
\label{app:sdvae}
The standard 1024-token latent is the f8c4 SD-VAE used by SiT
\citep{rombach2022ldm,ma2024sit}. The published SiT-XL/2 ImageNet-512
result reaches gFID $2.62$ after $3{,}000{,}000$ steps at batch size
$256$, using CFG $1.5$ and $250$ Euler--Maruyama steps
\citep{ma2024sit}. Under Equation~\ref{eq:app_flops} this is
$1.21\times10^{21}$ denoiser-training FLOPs.

We train SiT-XL/1 on APEX f64c$8{-}32{-}128$ with the optimizer in
Section~\ref{app:training_details}. At $1{,}000{,}000$ steps and batch
size $2{,}048$, this run reaches gFID $2.62$ using
$1.78\times10^{20}$ FLOPs, or $15\%$ of the f8c4 training compute.
The APEX evaluation uses CFG $1.5$ and $250$ Euler steps.

\begin{table}[ht]
\centering\small
\caption{Training compute to match the published f8c4 SiT-XL gFID on
ImageNet-512. Batch sizes are global. Training FLOPs use
Equation~\ref{eq:app_flops}.}
\label{tab:app_sdvae}
\begin{tabular*}{\linewidth}{@{}l@{\extracolsep{\fill}}rrrrr@{}}
\toprule
Latent & Tokens & Steps & Batch & Training FLOPs & gFID\\
\midrule
SD-VAE f8c4 & 1,024 & 3,000,000 & 256 & $1.21\times10^{21}$ & 2.62\\
APEX f64c$8{-}32{-}128$ & 64 & 1,000,000 & 2,048 & $1.78\times10^{20}$ & 2.62\\
\bottomrule
\end{tabular*}
\end{table}

\section{Autoencoder Training}
\label{app:ae_training}

We train all autoencoders on ImageNet \citep{deng2009imagenet} with a global batch size of 64. We follow
the DC-AE architecture, training recipe, and published implementation
\citep{chen2025dcae}, including its optimization settings and reconstruction,
perceptual \citep{zhang2018lpips}, and adversarial losses. APEX adds channel-prefix sampling and
resolution-aware supervision; the encoder and decoder remain shared across
all levels.

At each training step, we sample a cumulative level uniformly, retain its
channels, and set all later channels to zero. The decoder produces a
full-resolution image. We then resize both the reconstruction and the target
to the selected level's resolution before computing the loss.

\paragraph{Channel levels and target resolutions.}
Channel counts are cumulative: prefixes of 8, 32, and 128 channels add
8, 24, and 96 channels on the same spatial grid. We use the same
structure-to-detail heuristic as in Section~\ref{sec:method_ae}: at lower
widths, a roughly \(4\times\) increase in cumulative channels corresponds
to a \(2\times\) increase in target side length. At larger widths, we
introduce intermediate levels to avoid excessively large refinement groups
and treat this proportionality as a guideline. Exact assignments are given
in Table~\ref{tab:app_hierarchies}.

\par\medskip\noindent\begin{minipage}{\linewidth}
\centering\small\setlength{\tabcolsep}{3pt}
\captionof{table}{APEX hierarchies, target side lengths, and main SiT-XL/1 clock settings for ImageNet-512. The gap and training allocation are defined in Section~\ref{app:sampling}.}
\label{tab:app_hierarchies}
\label{tab:app_clock_config}
\begin{tabular}{@{}llllr@{}}
\toprule
Latent & Cumulative channels (group sizes) & Target side lengths & $g$ & $p_{L_0}$\\
\midrule
f64c128 & 8--32--128 \;(8, 24, 96) & 128, 256, 512 & 0.25 & 0.9\\
f64c256 & 8--32--128--256 \;(8, 24, 96, 128) & 128, 256, 512, 512 & 0.25 & 0.9\\
f64c512 & 8--32--128--256--512 \;(8, 24, 96, 128, 256) & 128, 256, 512, 512, 512 & 0.25 & 0.9\\
f128c512 & 32--128--256--512 \;(32, 96, 128, 256) & 128, 256, 512, 512 & 0.25 & 0.9\\
f128c1024 & 32--128--256--512--1024 \;(32, 96, 128, 256, 512) & 128, 128, 256, 256, 512 & 0.25 & 0.9\\
\bottomrule
\end{tabular}
\end{minipage}\par\medskip

\paragraph{Full-resolution control.}
The full-resolution control keeps the 8--32--128 channel hierarchy and prefix
sampling, but supervises every prefix at $512\times512$
(row~(e) of Table~\ref{tab:component_ablation}).

Algorithm~\ref{alg:apex_ae} shows the training loop, with APEX changes in green.
The base loss is the DC-AE objective for the current training stage.

\noindent\begin{minipage}{\linewidth}
\apexAlgCaption{Autoencoder training}{alg:apex_ae}
\noindent\begin{minipage}[t]{.48\linewidth}
\textbf{Standard autoencoder}
\begin{lstlisting}[style=apexpseudo]
for images in data:
  z = encoder(images)
  pred = decoder(z)
  loss = base_loss(pred, images)
  update(loss)
\end{lstlisting}
\end{minipage}\hfill
\begin{minipage}[t]{.49\linewidth}
\textbf{APEX}
\begin{lstlisting}[style=apexpseudo]
for images in data:
  z = encoder(images)
@@  level = uniform_level()@@
@@  z = keep_prefix(z, level)@@
  pred = decoder(z)
@@  size = target_size[level]@@
@@  pred = resize(pred, size)@@
@@  target = resize(images, size)@@
@@  loss = base_loss(pred, target)@@
  update(loss)
\end{lstlisting}
\end{minipage}
\end{minipage}

\texttt{keep\_prefix} zeros all channels after the sampled level, leaving the
decoder input at full width. The loss is the image-space objective on
\texttt{pred} and \texttt{target}; the zeroed channels are not supervised.

\paragraph{Reconstruction quality.}
\label{app:ae_metrics}
We train the f64c128 and f128c512 autoencoders to match the rFID of the
corresponding official DC-AE models.

\section{Staggered Training and Inference}
\label{app:sampling}
\label{app:diffusion_training}

\subsection{The staggered clock}
The global clock runs from $\tau=0$ to $T=1+(L-1)g$, where $L$ is the number
of levels and $g$ is the stagger gap. Each level has a local diffusion time
\begin{equation}
t_\ell(\tau)=\operatorname{clip}(1+\ell\cdot g-\tau,0,1).
\label{eq:app_clock}
\end{equation}
Local time decreases from 1 (noise) to 0 (clean). Level $\ell$ starts at
$\tau=\ell\cdot g$ (a multiple of the gap) and finishes at $\tau=1+\ell\cdot g$. Earlier levels are therefore
cleaner than the later levels they condition.

\begin{figure}[ht]
\centering
\begin{tikzpicture}
\begin{axis}[
 width=.94\linewidth,height=1.95in,xmin=0,xmax=1.5,ymin=-.04,ymax=1.08,
 xlabel={Global clock $\tau$},ylabel={Local timestep $t_\ell$},
 xtick={0,.25,.5,.75,1,1.25,1.5},ytick={0,.25,.5,.75,1},
 tick label style={font=\small},label style={font=\small},
 legend style={at={(.5,1.03)},anchor=south,legend columns=3,font=\small,draw=none},
 axis lines=left,grid=major,grid style={gray!18}]
\addplot[thick,mark=none] coordinates {(0,1)(1,0)(1.5,0)};
\addlegendentry{$L_0$: 8 channels}
\addplot[thick,dashed,mark=none] coordinates {(0,1)(.25,1)(1.25,0)(1.5,0)};
\addlegendentry{$L_1$: +24 channels}
\addplot[thick,dashdotted,mark=none] coordinates {(0,1)(.5,1)(1.5,0)};
\addlegendentry{$L_2$: +96 channels}
\end{axis}
\end{tikzpicture}
\caption{Three-level staggered clock with $g=0.25$. Before a level starts,
its input is zero. During the descending segment it is denoised. Once its local
time reaches zero, it stays fixed as context for the remaining levels.}
\label{fig:app_clock}
\end{figure}
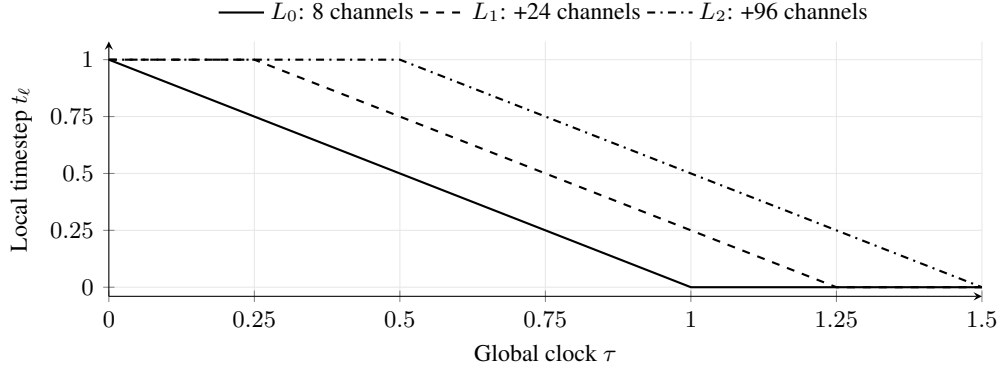

With $g=0.25$, the three denoising windows are $(0,1)$, $(0.25,1.25)$,
and $(0.5,1.5)$ (Figure~\ref{fig:app_clock}). Later levels begin before the
earlier levels have finished.

\subsection{Training-clock sampling}
We place probability $p_{L_0}$ on the interval $(0,1)$, where $L_0$ is still
denoising, and the remaining probability on $(1,T)$. Sampling is uniform within
each interval:
\begin{equation}
q(\tau)=\begin{cases}
p_{L_0}, & 0<\tau<1,\\
\dfrac{1-p_{L_0}}{T-1}, & 1<\tau<T.
\end{cases}
\label{eq:app_density}
\end{equation}
Uniform sampling over the full clock corresponds to $p_{L_0}=1/T$. When $g=0$,
the whole clock is $(0,1)$ and all levels share the same timestep.

Most training examples are drawn while $L_0$ is active because it must learn
structure generation. Refinement levels also receive training during the parts
of their trajectories that overlap this interval. For example, with three
levels, $g=0.25$, and $p_{L_0}=1$, $L_1$ is trained at local times $(0.25,1)$
and $L_2$ at $(0.5,1)$. Inference still traverses their complete trajectories.

Main-model clock settings are listed in Table~\ref{tab:app_clock_config}.

\subsection{Diffusion training}
Unstarted groups are zero, active groups are mixed with Gaussian noise, and
completed groups provide clean context. The target velocity is noise minus the
clean latent. We average the squared error within each active group and then
equally across active groups. Unstarted and completed groups do not contribute
to the loss.

The input-projection weights for $L_0$ use the standard SiT initialization \citep{ma2024sit};
those for later groups are initialized to zero. The Transformer is conditioned
on the global clock.

\noindent\begin{minipage}{\linewidth}
\apexAlgCaption{Diffusion training}{alg:apex_train}
\noindent\begin{minipage}[t]{.48\linewidth}
\textbf{Standard diffusion}
\begin{lstlisting}[style=apexpseudo]
for images, labels in data:
  z = encode(images)
  noise = gaussian_like(z)
  time = sample_time()
  noisy = mix(z, noise, time)
  pred = model(noisy, time, labels)
  target = noise - z
  loss = mse(pred, target)
  update(loss)
\end{lstlisting}
\end{minipage}\hfill
\begin{minipage}[t]{.49\linewidth}
\textbf{APEX}
\begin{lstlisting}[style=apexpseudo]
for images, labels in data:
  z = encode(images)
  noise = gaussian_like(z)
@@  clock = sample_clock()@@
@@  times = local_times(clock)@@
@@  noisy = prepare_groups(@@
@@      z, noise, times)@@
@@  pred = model(noisy, clock, labels)@@
  target = noise - z
@@  loss = active_group_mse(@@
@@      pred, target, times)@@
  update(loss)
\end{lstlisting}
\end{minipage}
\end{minipage}

\texttt{sample\_clock} draws $\tau\sim q$. \texttt{mix} forms $(1-t)z+t\epsilon$.
\texttt{prepare\_groups} sets the three group states described above;
\texttt{active\_group\_mse} averages their active losses. Class dropout is
unchanged.

\subsection{Diffusion inference}
Each group is initialized from Gaussian noise when its window opens. One
shared prediction updates all groups using their local-time changes:
\begin{equation}
z_\ell\leftarrow z_\ell+
\bigl[t_\ell(\tau_{k+1})-t_\ell(\tau_k)\bigr]v_\ell.
\label{eq:app_euler}
\end{equation}
Unstarted and completed groups have zero time change and remain fixed. The
grid includes the activation points $\ell\cdot g$ (multiples of the gap), where new groups are initialized
before the model evaluation.

\noindent\begin{minipage}{\linewidth}
\apexAlgCaption{Diffusion inference}{alg:sampling}
\noindent\begin{minipage}[t]{.48\linewidth}
\textbf{Standard diffusion}
\begin{lstlisting}[style=apexpseudo]
z = gaussian_latent()
for time, next_time in time_grid:
  pred = predict(z, time, label)
  delta = next_time - time
  z = z + delta * pred
image = decode(z)
\end{lstlisting}
\end{minipage}\hfill
\begin{minipage}[t]{.49\linewidth}
\textbf{APEX}
\begin{lstlisting}[style=apexpseudo]
@@z = zeros_latent()@@
@@for clock, next_clock in grid:@@
@@  start_new_groups(z, clock)@@
@@  pred = predict(z, clock, label)@@
@@  delta = local_times(next_clock)@@
@@  delta = delta - local_times(clock)@@
  z = z + delta * pred
image = decode(z)
\end{lstlisting}
\end{minipage}
\end{minipage}

\texttt{predict} combines the conditional and null-class predictions as
$v=v_{\mathrm{null}}+s(v_{\mathrm{class}}-v_{\mathrm{null}})$ when guidance is
used. Setting $s=1$ gives unguided class-conditional sampling.
\texttt{start\_new\_groups} initializes each group once. The completed latent
is decoded by the shared autoencoder.

\paragraph{Prefix evaluation.}
For Figure~\ref{fig:schedule_ablations}(c), we redistribute 250 NFEs over
$(0,1)$, $(0,1.25)$, or $(0,1.5)$ and decode the completed prefix, zeroing
later channels. These intervals complete one, two, or three levels at $g=0.25$.

\section{Complete Numerical Results}
\label{app:numerical}
\label{app:component_analysis}
\label{sec:model_size_scaling}
The tables below give the values in the main figures and additional generation
metrics. Each row contains one model or experimental setting. Unguided gFID is
class-conditional sampling at CFG 1.0. It is reported in
Tables~\ref{tab:app_channel_f64} and~\ref{tab:app_channel_f128} for every
setting with a matched guided run, and for the SiT-B gap and clock-allocation
ablations. A dash marks a setting whose guided run was not identified. sFID
and IS in Tables~\ref{tab:app_channel_f64} and~\ref{tab:app_channel_f128} are
computed on the same unguided samples. The remaining tables report guided
gFID, and their sFID and IS use the guided samples.

\subsection{Generation across channel counts}
\label{app:channel_metrics}

\noindent\begin{minipage}[t]{\linewidth}
\vspace{0pt}
\centering
\small
\setlength{\tabcolsep}{4pt}

\captionof{table}{
Figure~\ref{fig:channel_scaling}(a): generation at f64.
SiT-XL/1 models use 500K training steps and batch size 2,048.
Unguided gFID, sFID, and IS use CFG 1.0.
A dash denotes a metric that was not directly evaluated.
}
\label{tab:app_channel_f64}

\begin{tabular*}{\linewidth}{@{}r@{\extracolsep{\fill}}lrrrr@{}}
\toprule
& & \multicolumn{2}{c}{gFID $\downarrow$}
& sFID $\downarrow$ & IS $\uparrow$ \\
\cmidrule(lr){3-4}
Channels & Method & w/o CFG & CFG 1.5 & w/o CFG & w/o CFG \\
\midrule

128 & DC-AE
& 13.45 & 3.22 & 6.40 & 106 \\

128 & DC-AE 1.5
& 13.13 & 3.15 & 6.3 & 109 \\

128 & FreqWarm
& 15.61 & 3.53 & 6.92 & 100 \\

128 & Spectrum Matching
& 13.96 & 3.41 & 6.04 & 99 \\

128 & \textbf{APEX}
& \textbf{10.94} & \textbf{2.85}
& \textbf{6.05} & \textbf{121} \\

\midrule

256 & DC-AE
& 30.55 & 9.19 & 9.33 & 58 \\

256 & DC-AE 1.5
& 22.77 & 7.11 & 7.5 & 69 \\

256 & FreqWarm
& 24.96 & 7.54 & 7.5 & 64 \\

256 & Spectrum Matching
& 29.35 & 8.28 & 7.44 & 57 \\

256 & \textbf{APEX}
& \textbf{14.82} & \textbf{3.15}
& \textbf{7.42} & \textbf{108} \\

\midrule

512 & DC-AE
& 54.25 & 26.10 & 13.31 & 31 \\

512 & DC-AE 1.5
& 37.85 & 14.76 & 10.8 & 60 \\

512 & FreqWarm
& 46.02 & 20.80 & 12.1 & 45 \\

512 & Spectrum Matching
& 54.11 & 26.4 & 10.92 & 70 \\

512 & \textbf{APEX}
& \textbf{22.49} & \textbf{4.90}
& \textbf{8.57} & \textbf{86} \\

\bottomrule
\end{tabular*}
\end{minipage}

\par\medskip

\noindent\begin{minipage}[t]{\linewidth}
\vspace{0pt}
\centering
\small
\setlength{\tabcolsep}{4pt}

\captionof{table}{
Figure~\ref{fig:channel_scaling}(b): generation at f128.
SiT-XL/1 models use 1M training steps and batch size 4,096.
Unguided gFID, sFID, and IS use CFG 1.0.
}
\label{tab:app_channel_f128}

\begin{tabular*}{\linewidth}{@{}r@{\extracolsep{\fill}}lrrrr@{}}
\toprule
& & \multicolumn{2}{c}{gFID $\downarrow$}
& sFID $\downarrow$ & IS $\uparrow$ \\
\cmidrule(lr){3-4}
Channels & Method & w/o CFG & CFG 1.5 & w/o CFG & w/o CFG \\
\midrule

512 & DC-AE
& 41.01 & 14.44 & 11.91 & 49 \\

512 & DC-AE 1.5
& 36.96 & 11.86 & 12.55 & 57 \\

512 & FreqWarm
& 39.54 & 13.57 & 11.75 & 50 \\

512 & Spectrum Matching
& 40.63 & 13.77 & 12.95 & 51 \\

512 & \textbf{APEX}
& \textbf{31.01} & \textbf{8.98}
& \textbf{14.49} & \textbf{65} \\

\midrule

1024 & DC-AE
& 115.74 & 89.47 & 32.97 & 13 \\

1024 & DC-AE 1.5
& 84.92 & 51.47 & 28.12 & 26 \\

1024 & FreqWarm
& 109.09 & 79.51 & 28.15 & 14 \\

1024 & Spectrum Matching
& 107.74 & 73.71 & 34.33 & 15 \\

1024 & \textbf{APEX}
& \textbf{48.56} & \textbf{18.77}
& \textbf{18.98} & \textbf{42} \\

\bottomrule
\end{tabular*}
\end{minipage}

\par\medskip
\subsection{Schedule, component, and hierarchy ablations}
\label{app:ablation_metrics}
The Figure~\ref{fig:schedule_ablations} and component ablations use SiT-B/1 at
f64c128. Training FLOPs are matched within each comparison. Guided results
use CFG 1.5. Unguided gFID is reported for the gap and clock-allocation
comparisons.

\noindent\begin{minipage}[t]{.485\linewidth}\vspace{0pt}\centering\small\setlength{\tabcolsep}{2.5pt}
\renewcommand{\arraystretch}{1.08}
\captionsetup{justification=raggedright,singlelinecheck=false}
\captionof{table}{Figure~\ref{fig:schedule_ablations}(a): stagger gap. $\dagger$ extends the main plot.}
\label{tab:app_gap}
\begin{tabularx}{\linewidth}{@{}>{\raggedright\arraybackslash}Xrrrr@{}}
\toprule
 & \multicolumn{2}{c}{gFID $\downarrow$} & sFID$\downarrow$ & IS$\uparrow$\\
\cmidrule(lr){2-3}
Gap $g$ & \shortstack{w/o\\CFG} & \shortstack{CFG\\1.5} & & \\
\midrule
0.00 & 50.01 & 21.08 & 8.64 & 87 \\
0.10 & 40.10 & 13.73 & 7.75 & 120 \\
0.25 & 31.38 & 8.52 & 7.32 & 160 \\
0.50 & 33.48 & 9.35 & 10.00 & 166 \\
0.75$^{\dagger}$ & 45.40 & 16.66 & 14.34 & 130 \\
\bottomrule\end{tabularx}

\end{minipage}\hfill
\begin{minipage}[t]{.485\linewidth}\vspace{0pt}\centering\small\setlength{\tabcolsep}{2.5pt}
\renewcommand{\arraystretch}{1.08}
\captionsetup{justification=raggedright,singlelinecheck=false}
\captionof{table}{Figure~\ref{fig:schedule_ablations}(b): training allocation at $g=0.25$.}
\label{tab:app_allocation}
\begin{tabularx}{\linewidth}{@{}>{\raggedright\arraybackslash}Xrrrr@{}}
\toprule
 & \multicolumn{2}{c}{gFID $\downarrow$} & sFID$\downarrow$ & IS$\uparrow$\\
\cmidrule(lr){2-3}
$p_{L_0}$ & \shortstack{w/o\\CFG} & \shortstack{CFG\\1.5} & & \\
\midrule
$2/3$ & 35.45 & 10.87 & 7.09 & 137 \\
0.9 & 31.38 & 8.52 & 7.32 & 160 \\
1 & 30.77 & 8.12 & 6.78 & 163 \\
\bottomrule\end{tabularx}
\par\smallskip{\raggedright $p_{L_0}=2/3$ is uniform clock sampling.\par}

\end{minipage}
\par\bigskip
\noindent\begin{minipage}[t]{.485\linewidth}\vspace{0pt}\centering\small\setlength{\tabcolsep}{2.5pt}
\renewcommand{\arraystretch}{1.08}
\captionsetup{justification=raggedright,singlelinecheck=false}
\captionof{table}{Figure~\ref{fig:schedule_ablations}(c): completed prefixes of the $p_{L_0}=1$ model.}
\label{tab:app_prefix}
\begin{tabularx}{\linewidth}{@{}>{\raggedright\arraybackslash}Xrrr@{}}
\toprule
Channels & \shortstack{gFID $\downarrow$\\CFG 1.5} & sFID$\downarrow$ & IS$\uparrow$\\
\midrule
8 & 19.83 & 19.76 & 88 \\
32 & 9.04 & 8.89 & 150 \\
128 & 8.12 & 6.78 & 163 \\
\bottomrule\end{tabularx}
\par\smallskip{\raggedright The clock stops at 1, 1.25, and 1.5 for the first one, two, and three levels, respectively. Each row redistributes all 250 NFEs over that interval.\par}

\end{minipage}\hfill
\begin{minipage}[t]{.485\linewidth}\vspace{0pt}\centering\small\setlength{\tabcolsep}{2.5pt}
\renewcommand{\arraystretch}{1.08}
\captionsetup{justification=raggedright,singlelinecheck=false}
\captionof{table}{Additional metrics for the component study in Table~\ref{tab:component_ablation}.}
\label{tab:app_components}
\begin{tabularx}{\linewidth}{@{}>{\raggedright\arraybackslash}Xrrr@{}}
\toprule
Configuration & \shortstack{gFID $\downarrow$\\CFG 1.5} & sFID$\downarrow$ & IS$\uparrow$\\
\midrule
\shortstack[l]{(a) DC-AE /\\synchronous} & 13.27 & 7.42 & 117 \\
\addlinespace[2pt]
\shortstack[l]{(b) APEX AE /\\channel mean} & 20.37 & 7.69 & 86 \\
\addlinespace[2pt]
\shortstack[l]{(c) APEX AE /\\level mean} & 21.08 & 8.64 & 87 \\
\addlinespace[2pt]
\shortstack[l]{(d) APEX AE /\\prefix sampling} & 17.83 & 10.50 & 109 \\
\addlinespace[2pt]
\shortstack[l]{(e) Full-res. AE /\\stagger} & 8.77 & 6.87 & 161 \\
\addlinespace[2pt]
\shortstack[l]{(f) DC-AE 1.5 /\\stagger} & 52.89 & 11.48 & 31 \\
\addlinespace[2pt]
\shortstack[l]{(g) Full APEX} & 8.12 & 6.78 & 163 \\
\addlinespace[2pt]
\bottomrule\end{tabularx}

\end{minipage}
\par\medskip

\section{Figure Reproduction}
\label{app:reproduction}
\label{app:figure_reproduction}
This section describes how we produce each figure. Training and evaluation
follow Section~\ref{app:training_details}; channel hierarchies and the
staggered process are given in Sections~\ref{app:ae_training}
and~\ref{app:sampling}.

\subsection{Figure 1: overview}
\paragraph{Purpose.}
Show how APEX organizes a wide latent, denoises its levels asynchronously,
and changes the reconstruction--generation trade-off.
\paragraph{Procedure.}
Panel (a) is a schematic of an f128c512 latent: a \(512^2\) image is encoded
onto a \(4\times4\) grid of 16 tokens, and the 512 channels of each token are
ordered from structure to detail. Panel (b) shows the corresponding staggered
schedule on one global clock \(\tau\): each later level starts after a delay
\(g\), so it is denoised while earlier levels are already cleaner. Panel (c)
is a schematic of the reconstruction--generation frontier relative to a flat
latent; the points and curves are illustrative, not measured. The quantitative
five-method comparison at f128c512 is Figure~\ref{fig:channel_scaling}.

\subsection{Figure 2: autoencoder, training, and inference}
\paragraph{Purpose.}
Show how APEX changes autoencoder training, diffusion training, and sampling.
\paragraph{Procedure.}
The autoencoder panel encodes an image, samples a cumulative prefix, zeros
the remaining channels, and decodes at full resolution. Reconstruction and
target are then resized to the prefix's target resolution before the loss.
The example retains the first two of four levels.
The training panel samples a global clock from $q(\tau)$ and maps it to
local times with Equation~\ref{eq:app_clock}. With $g=0.25$ and $\tau=0.38$,
the first two local times are 0.62 and 0.87; the third level has not started.
Active groups are noised at these times, the unstarted group is zeroed, and
one shared Transformer predicts their velocities.
The inference panel shows the same three-level clock at the activation
points: at $\tau=0$ only $L_0$ is active; at $\tau=g$ and $\tau=2g$, $L_1$
and $L_2$ are initialized from Gaussian noise. Generation continues until
every group is clean.

\subsection{Figure 3: staged generation}
\paragraph{Purpose.}
Show the image after each level reaches $t=0$.
\paragraph{Procedure.}
We run the three-level f64c8--32--128 model and decode the completed
prefix when each successive level finishes. The displayed prefixes
contain 8, 32, and 128 channels. Within each example, all three images
come from the same generation, and the inset shows the same crop.
The difference visualizations in Section~\ref{app:qual_differences}
subtract successive decodes of these two examples.

\subsection{Figure 4: training-clock density}
\paragraph{Purpose.}
Show how training examples are allocated along the global clock.
\paragraph{Procedure.}
We use a three-level clock with $g=0.25$, so its endpoint is $T=1.5$.
Uniform clock sampling has density $1/T=2/3$ throughout the interval.
For the structure-heavy example, we assign probability 0.9 to $(0,1)$,
where $L_0$ is active, and probability 0.1 to $(1,1.5)$. Sampling uniformly
within each interval gives density 0.9 in the first interval and 0.2 in the
second. The figure plots these two densities; it does not require training
another model.

\subsection{Figure 5: generation across channel counts}
\paragraph{Purpose.}
Compare generation as the latent becomes wider at fixed spatial compression
and training compute.
\paragraph{Procedure.}
We compare DC-AE, DC-AE 1.5, FreqWarm, Spectrum Matching, and APEX at
f64c128, f64c256, f64c512, f128c512, and f128c1024. Autoencoders are
trained on ImageNet with the corresponding method's representation objective.
For APEX, full-latent reconstruction quality is matched to the corresponding
DC-AE model. We then train a SiT-XL/1 denoiser for each method and latent
shape. The f64 runs use 500K steps at batch size 2,048, and the f128 runs
use 1M steps at batch size 4,096, giving approximately the same cumulative
training FLOPs.

We generate 50,000 class-conditional images per model with CFG 1.5 and
compute gFID using the ADM protocol. Within each panel, the spatial grid
stays fixed and only the channel count changes. The f64 panel uses 64 tokens
and the f128 panel uses 16. Tables~\ref{tab:app_channel_f64}
and~\ref{tab:app_channel_f128} give the numerical values, including unguided
gFID at CFG 1.0 for every setting whose guided checkpoint was identified.

\subsection{Figure 6: inference compute and guidance}
\paragraph{Purpose.}
Test whether stronger spatial compression improves generation when both
training and inference compute are limited.
\paragraph{Procedure.}
We train DC-AE f32c32 for 250K steps at batch size 1,024, and DC-AE f64c128
and APEX f64c8--32--128 for 500K steps at batch size 2,048. These schedules
match denoiser-training compute. We then keep each trained model fixed and
change its number of sampling steps. Every NFE and guidance setting is
evaluated using 50,000 generated images.

For panel (a), we plot guided gFID against relative denoiser inference FLOPs.
One f32 evaluation costs approximately four f64 evaluations, so the
$1\times$ budget pairs 64 f32 NFEs with 250 f64 NFEs. The f32 curve uses
16, 32, 64, 128, and 250 NFEs. The f64 curves use 32, 64, 128, 250, 500, and 1,000 NFEs.
At each inference budget, we sweep CFG in increments of 0.1 independently
for each model, extending the sweep until gFID increases on both sides of
the minimum. We report the best evaluated value.

For panel (b), we hold inference compute at $1\times$ and plot gFID against
the guidance scale. At this budget, the resulting sweeps span 1.2--1.5 for
f32 and 1.3--1.6 for the f64 models. APEX and f32 select 1.4, while
DC-AE f64 selects 1.5.

\subsection{Figure 7: prefix reconstructions}
\paragraph{Purpose.}
Show the information retained by early prefixes.
\paragraph{Procedure.}
We encode the same image with DC-AE 1.5 and APEX. We retain a cumulative
prefix, set all later channels to zero, and decode at full resolution.
The DC-AE 1.5 examples use 16, 64, and 256 channels; APEX uses the
resolution-aware hierarchy shown in the figure. We display the same image
region at each level to compare the detail already present in the early
prefixes. The 16-channel insets use the same crop on both rows.

\subsection{Figure 8(a): stagger gap}
\paragraph{Purpose.}
Measure the trade-off between cleaner conditioning and overlap between
levels' denoising windows.
\paragraph{Procedure.}
We keep the resolution-aware f64c8--32--128 autoencoder, SiT-B/1 backbone,
training-clock allocation, and training FLOPs fixed, and train with
$g\in\{0,0.1,0.25,0.5\}$. At $g=0$, every group has the same timestep, while
the loss remains equally weighted across groups. For each gap, we sample
over the complete global clock with the same NFE budget and CFG 1.5, then
compute gFID from 50,000 images. Thus a larger gap changes the overlap and
clock length without increasing the sampling budget.
Table~\ref{tab:app_gap} also reports the $g=0.75$ extension.

\subsection{Figure 8(b): training allocation}
\paragraph{Purpose.}
Measure how concentrating training on structure generation affects quality.
\paragraph{Procedure.}
We keep the same three-level autoencoder and fix $g=0.25$. We train three
SiT-B/1 models at the same compute budget, changing only the clock-sampling
distribution: uniform over the full clock, $p_{L_0}=0.9$, and $p_{L_0}=1$.
For uniform sampling, $p_{L_0}=2/3$. The other two settings sample within
the intervals using Equation~\ref{eq:app_density}. All three models are
evaluated with the same sampling budget, CFG 1.5, and 50,000 generated images.

\Needspace{11\baselineskip}
\subsection{Figure 8(c): contribution of successive levels}
\paragraph{Purpose.}
Test whether later levels improve generation beyond the structural prefix.
\paragraph{Procedure.}
We use the same $p_{L_0}=1$ model for all three conditions. We stop the global
clock at 1, 1.25, or 1.5, when the first one, two, or three levels have
finished. In every case, we redistribute all 250 NFEs over the retained
interval. We then zero the channels after the completed prefix and decode
its 8, 32, or 128 channels with the same autoencoder. Each condition produces
50,000 images with CFG 1.5. The resulting gFIDs are 19.83, 9.04, and 8.12.
No additional diffusion models are trained for this comparison.

\subsection{Table 2: component analysis}
\paragraph{Purpose.}
Separate the effects of the representation, loss, and asynchronous training.
\paragraph{Procedure.}
We train all seven variants with SiT-B/1 at f64c128 and the same cumulative
training FLOPs. Row (a) uses the flat DC-AE latent, one shared timestep, and
a loss averaged over all channels. Row (b) applies the same training procedure
to the APEX autoencoder. Row (c) keeps the full latent and shared timestep,
but averages the loss within each channel group and then equally across
groups. Row (d) instead samples a cumulative prefix uniformly, zeros the
suffix, and noises the retained groups at one shared timestep; its loss is
averaged equally over those groups.

Row (e) keeps the 8--32--128 channel partition and APEX diffusion, but trains every
autoencoder prefix against the full-resolution target. Row (f) uses the
DC-AE 1.5 hierarchy with 16, 20, $\ldots$, 128 channels and uniform cumulative-prefix
sampling, with staggered local times on the retained groups. Row (g) uses
the complete resolution-aware APEX hierarchy and diffusion procedure. We
evaluate each model using the common generation protocol.

\subsection{Additional visualizations and comparisons}
\paragraph{Purpose.}
Show progressive generation and refinement differences.
\paragraph{Procedure.}
For Figure~\ref{fig:app_generations}, we sample from the staggered generator:
eight examples with f64c8--32--128, four with f64c8--32--128--256, and four
with f64c8--32--128--256--512. When each successive level reaches $t=0$, we
decode the completed prefix with the shared autoencoder. These are generations,
not reconstructions of real images.

For Figure~\ref{fig:app_differences}, we subtract consecutive full-resolution
prefix decodes before resizing or adding the inset crops. Both examples and
both differences use the same display scale.

Figure~\ref{fig:app_clock} evaluates Equation~\ref{eq:app_clock} for three
levels and $g=0.25$.

\end{document}